\documentclass{article}

\usepackage{arxiv}
\usepackage[utf8]{inputenc} %
\usepackage[T1]{fontenc}    %
\usepackage{microtype}      %
\usepackage{amsfonts}
\usepackage{amsmath}
\usepackage{bm}
\usepackage[hidelinks]{hyperref}
\usepackage{graphicx}
\usepackage{mathtools}
\usepackage{enumitem}
\usepackage{setspace}
\usepackage{subcaption}
\usepackage[dvipsnames]{xcolor}
\usepackage{color}
\graphicspath{{figures/}}
\usepackage{upgreek}
\usepackage{cases}
\usepackage{arydshln}
\usepackage{wrapfig}
\usepackage{blkarray}
\usepackage{enumitem}
\usepackage{textcomp}  %
\usepackage{gensymb}  %
\usepackage{siunitx}  %
\usepackage{booktabs}
\usepackage{longtable}
\usepackage[export]{adjustbox}%
\usepackage[ruled,vlined]{algorithm2e}

\newcommand{\ie}{\textit{i.e.}}
\newcommand{\eg}{\textit{e.g.}}

\newcommand{\figref}[1]{Figure~\ref{fig:#1}}
\newcommand{\tabref}[1]{Table~\ref{tab:#1}}
\newcommand{\secref}[1]{\S\ref{sec:#1}}
\newcommand{\apxref}[1]{Appendix~\ref{apx:#1}}
\newcommand{\eqnref}[1]{\eqref{eqn:#1}}
\DeclareMathOperator*{\argmin}{arg\,min}

\newcommand\R{\mathbb{R}}

\newcommand\T{\mathbf{T}}
\newcommand\Or[0]{\mathrm{\mathbf{O}}}
\newcommand\SO[0]{\mathrm{\mathbf{SO}}}

\newcommand\x{\mathbf{x}}
\newcommand\p{\mathbf{p}}
\newcommand\f{\mathbf{f}}
\newcommand\G{\mathcal{G}}

\newcommand\bth[0]{{\boldsymbol{\theta}}}

\usepackage{tikz,pgfplots,pgfplotstable}
\usepgfplotslibrary{groupplots,fillbetween,colorbrewer,statistics}
\usetikzlibrary{intersections,calc,arrows,matrix,spy,pgfplots.statistics, pgfplots.colorbrewer}
\usepackage{color}
\definecolor{darkred}{rgb}{0.7,0,0}
\definecolor{darkgreen}{rgb}{0,0.5,0}
\definecolor{darkblue}{rgb}{0,0,0.7}
\definecolor{SkyBlue}{rgb}{0.53, 0.81, 0.92}
\pgfplotsset{compat=1.5.1, cycle list/Set1-3}
\usepackage{tikz-3dplot}

\author{
    \href{https://orcid.org/0000-0001-7373-4150}{\includegraphics[scale=0.06]{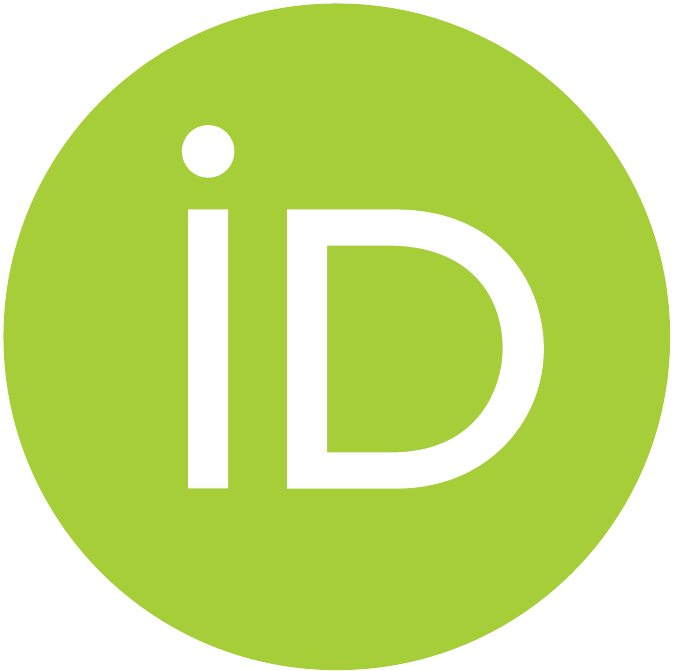}\hspace{1mm}}Jelena Banjac,
    \href{https://orcid.org/0000-0001-9834-7755}{\includegraphics[scale=0.06]{orcid}\hspace{1mm}}Laurène Donati,
    \href{https://orcid.org/0000-0002-6028-9024}{\includegraphics[scale=0.06]{orcid}\hspace{1mm}}Michaël Defferrard \\
    EPFL, Switzerland \\
    \texttt{\{jelena.banjac,laurene.donati,michael.defferrard\}@epfl.ch}
}

\date{\today}

\renewcommand{\headeright}{}  %
\renewcommand{\undertitle}{}  %
\title{Learning to recover orientations from\\projections in single-particle cryo-EM}
\begin{document}

\maketitle

\begin{abstract}
    A major challenge in single-particle cryo-electron microscopy (cryo-EM) is that the orientations adopted by the 3D particles prior to imaging are unknown; yet, this knowledge is essential for high-resolution reconstruction.
    We present a method to recover these orientations directly from the acquired set of 2D projections.
    Our approach consists of two steps: (i) the estimation of distances between pairs of projections, and (ii) the recovery of the orientation of each projection from these distances.
    In step (i), pairwise distances are estimated by a Siamese neural network trained on synthetic cryo-EM projections from resolved bio-structures.
    In step (ii), orientations are recovered by minimizing the difference between the distances estimated from the projections and the distances induced by the recovered orientations.
    We evaluated the method on synthetic cryo-EM datasets.
    Current results demonstrate that orientations can be accurately recovered from projections that are shifted and corrupted with a high level of noise.
    The accuracy of the recovery depends on the accuracy of the distance estimator.
    While not yet deployed in a real experimental setup, the proposed method offers a novel learning-based take on orientation recovery in SPA.
    Our code is available at \url{https://github.com/JelenaBanjac/protein-reconstruction}.
\end{abstract}

\section{Introduction}\label{sec:introduction}

Single-particle cryo-electron microscopy (cryo-EM) has revolutionized the field of structural biology over the last decades~\cite{dubochet1988cryo, frank2006three, chap0-nat2015MethodYear}.
The use of electron beams to image ice-embedded samples has permitted the recovery of 3D bio-structures at unprecedented resolution.
This ``resolution revolution'' has had a tremendous impact in biomedical research, providing invaluable insights into the biological processes that underlie many current diseases.

\begin{figure}
    \begin{minipage}[t]{0.48\linewidth}
        \centering
        \includegraphics[height=0.7\linewidth]{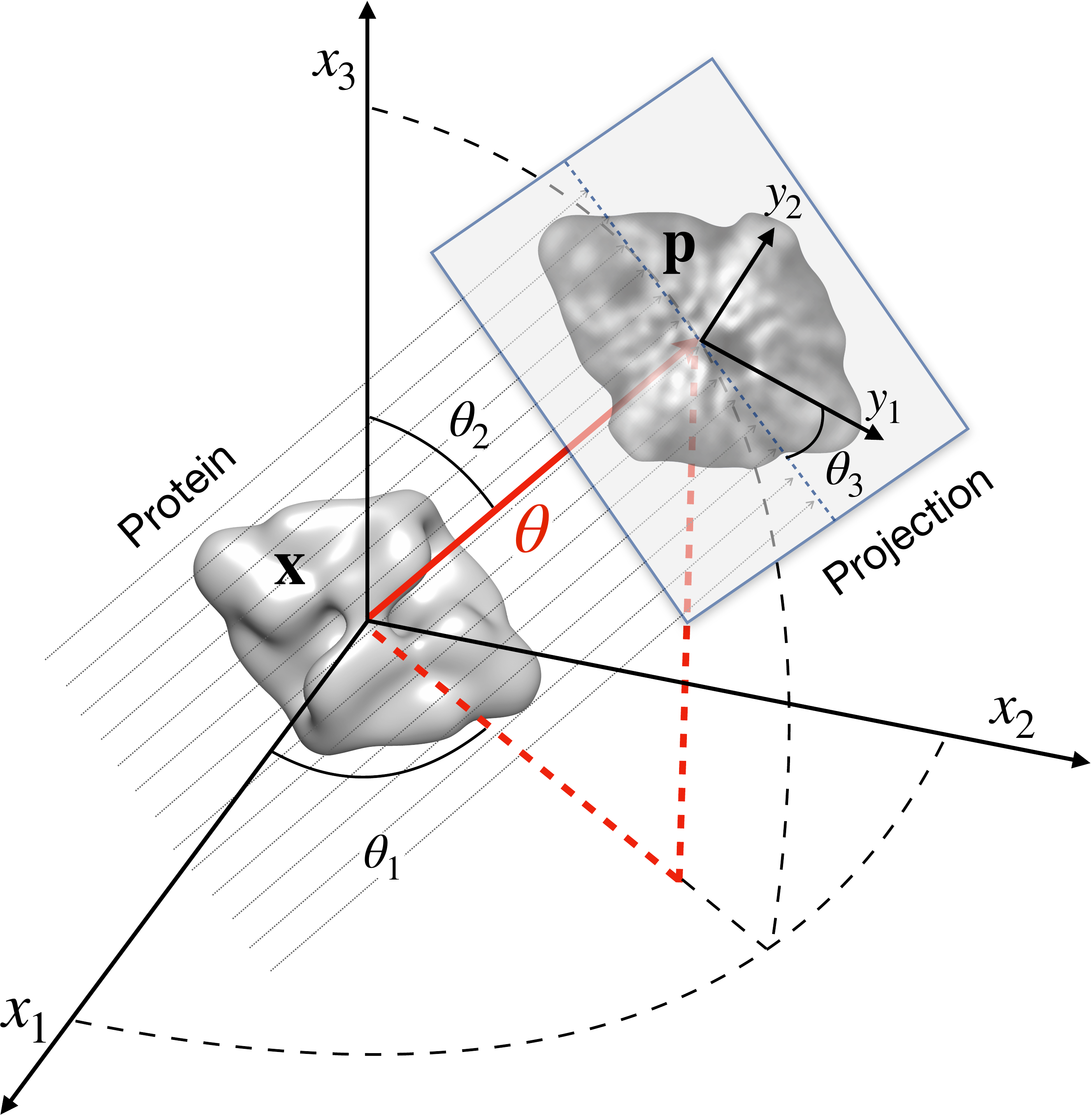}
        \caption{%
            Geometry of the imaging model defined in \eqnref{imaging-model}.
            The 3D density $\x$ in the coordinate system $(x_1, x_2, x_3)$ is imaged along the \textit{orientation} $\bth$ to produce the 2D \textit{projection} $\p$ in the coordinate system $(y_1, y_2)$ of the microscope's detector plane.
            The orientation $\bth = (\theta_3, \theta_2, \theta_1)$ is decomposed as the direction $(\theta_2, \theta_1) \in [0,\pi] \times [0,2\pi[$ (parameterizing the sphere $\mathbb{S}^2$) and the in-plane rotation $\theta_3 \in [0,2\pi[$ (parameterizing the circle $\mathbb{S}^1$).
            In our work, we represent the orientation $\bth$ as a unit quaternion $q$.
        }\label{fig:imaging-geometry}
    \end{minipage}
    \hfill
    \begin{minipage}[t]{0.48\linewidth}
        \centering
        \includegraphics[height=0.7\linewidth]{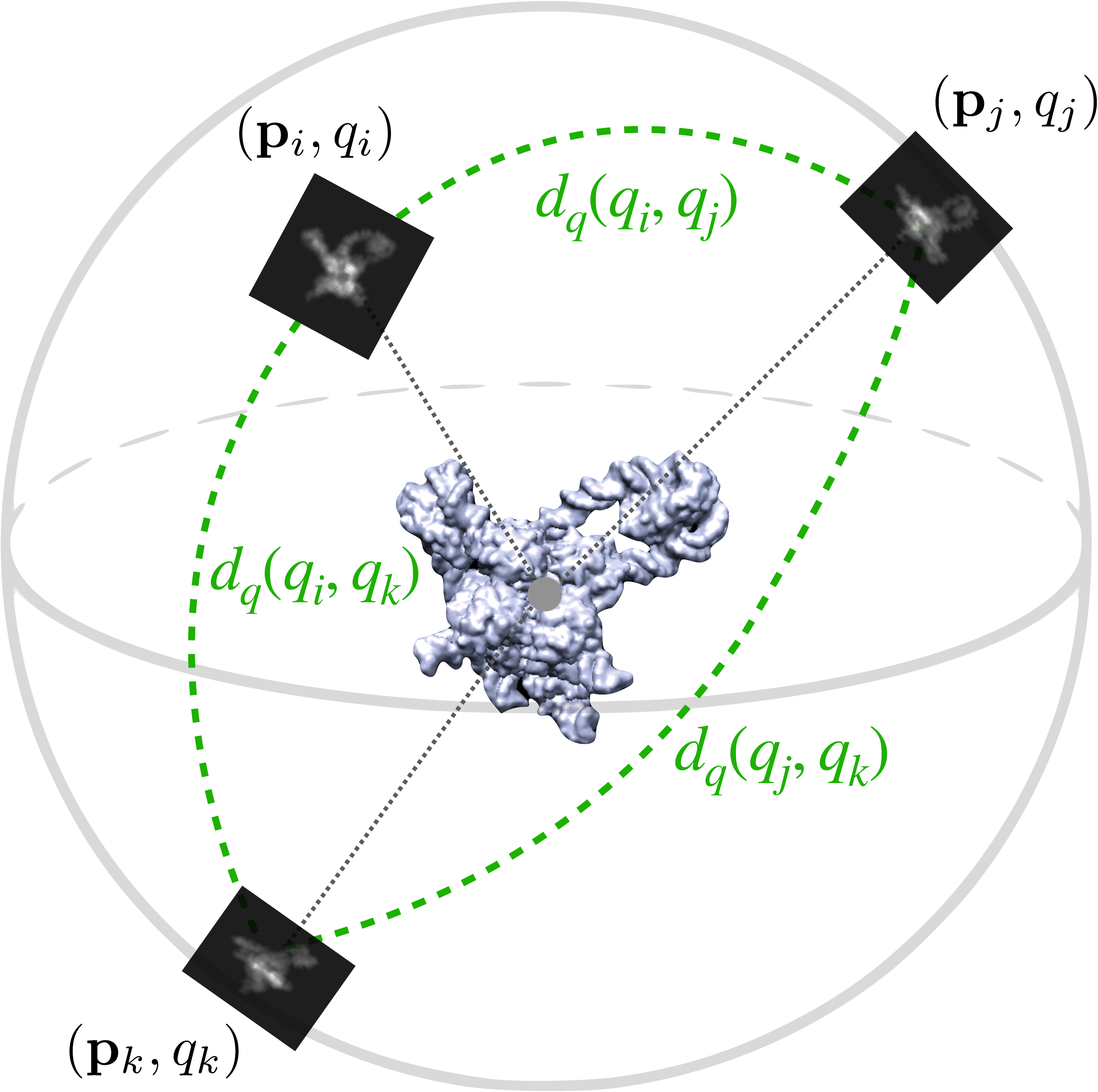}
        \caption{%
            Single-particle cryo-EM produces $P$ projections (with $P$ in the order of $10^5$) from unknown orientations: $\{(\p_i, q_i)\}_{i=1}^P$.
            Observing that distances between orientations constrain the latter, we aim to \textit{recover the orientations} $\{q_i\}$ from $\{d_q(q_i, q_j)\}$, where $d_q(q_i, q_j)$ is the distance (angle) between orientations $q_i$ and $q_j$.
            Observing that the similarity between projections depends on their relative orientation, we aim to \textit{estimate the distance} $d_q(q_i, q_j)$ from the projections $(\p_i, \p_j)$.
        }\label{fig:intuition-method}
    \end{minipage}
\end{figure}

In single-particle cryo-EM, every 3D particle adopts a random orientation $\bth_i$ in the ice layer before being imaged with parallel beams of electrons.
Hence, the projection geometry associated to each acquired 2D projection (\figref{imaging-geometry}) is unknown.
Yet, this knowledge is essential for the tomographic reconstruction of bio-structures~\cite{Natterer2001mathematics}.
We consider that a cryo-EM measurement (\ie, a projection) $\p_i \in \R^{n_p}$ is acquired through
\begin{equation}
    \label{eqn:imaging-model}
    \p_i = \mathbf{C}_{\boldsymbol\varphi} \mathbf{S}_{\mathbf{t}_i} \mathbf{P}_{\bth_i} \x + \mathbf{n},
\end{equation}
where $\x \in \R^{n_x}$ is the unknown 3D density map~\cite{dimaio_creating_2007} (Coulomb potential).
The operator $\mathbf{P}_{\bth_i}: \R^{n_x} \to \R^{n_p}$ is the projection along the orientation $\bth_i$ (\ie, the x-ray transform).
The operator $\mathbf{S}_{\mathbf{t}_i}: \R^{n_p} \to \R^{n_p}$ is a shift of the projection by $\mathbf{t}_i = (t_{i_1}, t_{i_2})$.
The convolution operator $\mathbf{C}_{\boldsymbol\varphi}: \R^{n_p} \to \R^{n_p}$ models the microscope point-spread function (PSF) with parameters $\boldsymbol\varphi = (d_1, d_2, \alpha_\mathrm{ast})$, where $d_1$ is the defocus-major, $d_2$ is the defocus-minor, and $\alpha_\mathrm{ast}$ is the angle of astigmatism~\cite{vulovic_image_2013,rullgard_simulation_2011}.
Finally, $\mathbf{n} \in \R^{n_p}$ represents additive noise.
\figref{different-projections} illustrates the effect of projection, shift, and noise.
The challenge is then to reconstruct $\x$ from a set of projections $\{\p_i\}_{i=1}^P$ acquired along unknown orientations.

A popular approach is to alternatively refine the 3D structure and an estimation of the orientations~\cite{penczek1994ribosome,Baker1996,Dempster1977,sigworth1998maximum,scheres2012bayesian,zehni2020joint}.
Yet, the outcome of these iterative-refinement procedures is still often predicated on the quality of the initial reconstruction, or, equivalently, on the initial estimation of the orientations~\cite{sorzano2006optimization,henderson2012outcome}.

Several methods have been designed to produce a first rough \textit{ab initio} structure for the refinement procedure~\cite{singer2020computational}.
An early approach~\cite{kam1980reconstruction} proposed to reconstruct an initial structure such that the first few moments of the distribution of its theoretical measurements match the ones of its experimental projections.
Since then, \textit{moment-matching} techniques have been refined and extended~\cite{salzman1990method,goncharov1988integral,sharon2019method}, \eg, to accommodate for non-uniform orientation configurations.
However, they typically remain sensitive to error in data and can require relatively high computational complexity.

Another popular approach relies on the central-slice theorem, which relates the Fourier transform of a projection to a plane (orthogonal to the projection direction) in the Fourier transform of the 3D object~\cite{Natterer2001mathematics}.
Hence, every two projections \textit{de facto} share a common 1D intersection in the 3D Fourier domain, and three projections theoretically suffice to define a coordinate system from which their orientations can be deduced~\cite{van1987angular}.
Exploiting this principle, \textit{common-lines} methods aim at uniquely determining the orientations of each projection by identifying the common-lines between triplets of projections~\cite{penczek1994ribosome,mallick2006structure,singer2010detecting,wang2013orientation,greenberg2017common,pragier2019common}---a real technical challenge given the massive amount of noise in cryo-EM data.

Alternatively, the marginalized maximum likelihood (ML) formulation of the reconstruction problem~\cite{sigworth1998maximum}---classically used for the iterative-refinement procedures themselves---can be minimized using stochastic gradient descent~\cite{punjani2017cryosparc}.
This permits to avoid the need for an initial volume estimate, at the possible cost of greater convergence instability.
More recently, the recovery of geometrical information from unknown view tomography of 2D point sources has been proposed~\cite{zehni2019distance}, but the extension to 3D cryo-EM tomography is not straightforward.
Finally, \cite{miolane2019estimation} proposed to recover the in-plane rotations by learning to embed projections in an appropriate latent space, but only after directions had been estimated through three rounds of 2D classification in RELION\@.

Despite the many aforementioned advances, the task of providing a robust initial volume remains an arduous challenge in single-particle cryo-EM due to the high-dimensionality and strong ill-posedness of the underlying optimization problem.
On the other hand, deep learning has had a profound influence in imaging in reason of the remarkable ability of convolutional neural networks to capture relevant representations of images~\cite{lecun2015deep}.

In this work, we present a new learning-based approach to recover the unknown orientations directly from the acquired set of projections.
By doing so, orientations are recovered without the need for an intermediate reconstruction procedure or an initial volume estimate. %

\section{Method}

\begin{figure}
    \centering
    \includegraphics[width=\linewidth]{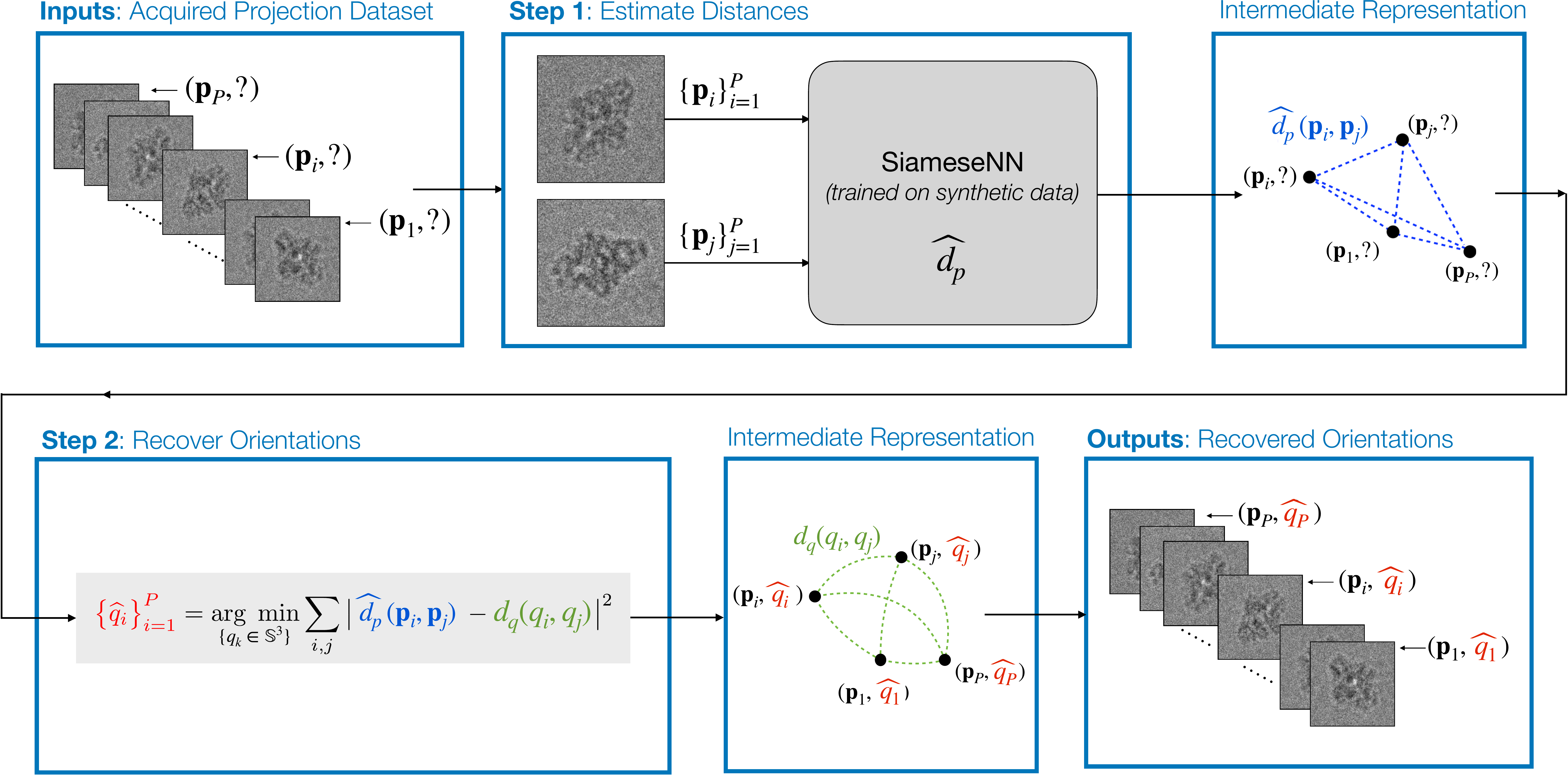}
    \caption{%
        Our method consists of two steps.
        First, we estimate distances between pairs of projections.
        Second, we recover the orientation of each projection from these distances.
    }\label{fig:schematic:method-overview}
\end{figure}

Our approach relies on two observations (\figref{intuition-method}), yielding two steps (\figref{schematic:method-overview}).
First, the greater the similarity between two 2D projections $(\p_i, \p_j)$, the more likely they originated from two 3D particles that adopted close orientations $(q_i, q_j)$ in the ice layer prior to imaging;\footnote{Up to protein symmetries, which we discuss later.} this observation guides a number of applications in the field~\cite{frank2006three}.
Hence, we aim to \textit{estimate distances} between orientations $d_q(q_i, q_j)$ from the projections themselves as $\widehat{d_p}(\p_i, \p_j)$, which we discuss in \secref{method:distance-learning}.
Second, an orientation $q$ is constrained by the distances between itself and the other orientations $\{ d(q, q_j) \}$.
Hence, we aim to \textit{recover orientations} $\{\widehat{q_k}\}$ such that the induced distances $\{ d_q(\widehat{q_i}, \widehat{q_j}) \}$ are close to the estimated distances $\{ \widehat{d_p}(\p_i, \p_j) \}$, which we discuss in \secref{method:orientation-recovery}.
All in all, from a set of projections $\{\p_k\}$, we aim to recover their orientations $\{\widehat{q_k}\}$ such that $d_q(\widehat{q_i}, \widehat{q_j}) \approx \widehat{d_p}(\p_i, \p_j) \approx d_q(q_i, q_j)$, with equality if $\widehat{d_p}$ and $\{\widehat{q_k}\}$ are perfectly estimated.
Our approach is similar to~\cite{coifman2008graphprojections}.
While the authors are there concerned with the reconstruction of 2D images from 1D projections, they rely on the same two-step approach: they (i) estimate distances as $\widehat{d_p}(\p_i, \p_j) = \| \p_i - \p_j \|_2$ then (ii) recover the orientations by spectrally embedding that distance graph.
The Euclidean distance is however not robust to perturbations:
for example, two projections that only differ by a shift $\mathbf{S_t}$ of one pixel would be considered far apart while their orientations are the same.
They noted that issue and we observed it too (\apxref{results:distance-estimation:euclidean}). To circumvent this, we propose to \textit{learn} $\widehat{d_p}$ from examples (\secref{method:distance-learning}).
\subsection{Representation of orientations with quaternions}\label{sec:method:orientation-representation}
The orientation of a 3D particle with respect to the microscope's detector plane is a rotation relative to a reference orientation (\figref{imaging-geometry}).
The group of all 3D rotations under composition is identified with $\SO(3)$, the group of $3 \times 3$ orthogonal matrices with determinant~1 under matrix multiplication.
A rotation matrix $\mathbf{R}_{\bth} \in \SO(3)$ can be decomposed as a product of $\binom{3}{2}=3$ independent rotations, for example as $\mathbf{R}_{\bth} = \mathbf{R}_{\theta_3} \mathbf{R}_{\theta_2} \mathbf{R}_{\theta_1}$, where $\bth = (\theta_3,\theta_2,\theta_1) \in [0,2\pi[ \, \times \, [0,\pi] \times [0,2\pi[$ are the (extrinsic and proper) Euler angles in the $ZYZ$ convention (a commonly-used parameterization in cryo-EM)~\cite{sorzano2014interchanging}.

While Euler angles are a concise representation of orientation ($3$ numbers for $3$ degrees of freedom), they suffer from a topological constraint---there is no covering map from the $3$-torus to $\SO(3)$---which manifests itself in the \textit{gimbal lock}, the loss of one degree of freedom when $\theta_2=0$. %
This makes their optimization by gradient descent (\secref{method:orientation-recovery}) problematic.
On the other hand, the optimization of rotation matrices (made of $9$ numbers) would require computationally costly constraints (orthogonality and determinant~1) to reduce the number of degrees of freedom to $3$.
Moreover, the distance between orientations cannot be directly computed from Euler angles and is costly (30 multiplications) to compute from rotation matrices~\cite{huynh2009metrics}.
We solve both problems by representing orientations with unit quaternions.

Quaternions $q \in \mathbb{H}$ are an extension of complex numbers\footnote{The algebra $\mathbb{H}$ is similar to the algebra of complex numbers $\mathbb{C}$, with the exception of multiplication being non-commutative.} of the form $q = a + b\boldsymbol{i} + c\boldsymbol{j} + d\boldsymbol{k}$ where $a,b,c,d \in \R$.
Unit quaternions $q \in \mathbb{S}^3$, where $\mathbb{S}^3 = \big\{ q \in \mathbb{H}: \lvert q \rvert = 1 \big\}$ is the 3-sphere (with the additional group structure inherited from quaternion multiplication), concisely and elegantly represent a rotation of angle $\theta$ about axis $(x_1, x_2, x_3)$ as $q = \cos(\theta/2) + x_1 \sin(\theta/2) \boldsymbol{i} + x_2 \sin(\theta/2) \boldsymbol{j} + x_3 \sin(\theta/2) \boldsymbol{k}$.
They parameterize rotation matrices as
\begin{equation*}
    \mathbf{R}_q =
    \begin{pmatrix}
        a^2+b^2-c^2-d^2 & 2bc-2ad & 2bd+2ac \\
        2bc+2ad & a^2-b^2+c^2-d^2 & 2cd-2ab \\
        2bd-2ac & 2cd+2ab & a^2-b^2-c^2+d^2
    \end{pmatrix}.
\end{equation*}
Note that $\mathbb{S}^3 \rightarrow \SO(3)$ is a two-to-one mapping (a double cover) as $q$ and $-q$ represent the same orientation.
Unlike Euler angles, $\mathbb{S}^3$ is isomorphic to the universal cover of $\SO(3)$.
Hence, the distance between two orientations, \ie, the length of the geodesic between them on $\SO(3)$, is given by
\begin{equation}
    \begin{aligned}
        d_q &: \mathbb{S}^3 \times \mathbb{S}^3 \rightarrow [0,\pi], \\
        d_q(q_i, q_j) &= 2 \arccos \left( \left| \langle q_i, q_j \rangle \right| \right),
    \label{eqn:distance:orientations}
    \end{aligned}
\end{equation}
where $\langle \cdot, \cdot \rangle$ is the inner product, and the absolute value $\left| \cdot \right|$ ensures that $d_q(q_i, q_j) = d_q(q_i, -q_j)$.
The distance $d_q(q_i, q_j)$ corresponds to the magnitude (angle $\theta$) of the rotation $\mathbf{R}_*$ such that $\mathbf{R}_{q_i} = \mathbf{R}_* \mathbf{R}_{q_j}$~\cite{huynh2009metrics}.

\subsection{Distance learning}\label{sec:method:distance-learning}

We aim to estimate a function $\widehat{d_p}$ such that $\widehat{d_p}(\p_i, \p_j) \approx d_q(q_i, q_j)$.
While we could in principle design $\widehat{d_p}$, that would be intricate---if not impossible---partly because the invariants are difficult to specify.
We instead opt to learn $\widehat{d_p}$, capitalizing on (i) the powerful function approximation capabilities of neural networks, and (ii) the possibility to generate realistic cryo-EM projection datasets supported by the availability of numerous 3D atomic models\footnote{\url{https://www.ebi.ac.uk/pdbe/emdb}} and our ability to model the imaging procedure.

\begin{figure}
    \centering
    \includegraphics[width=\linewidth]{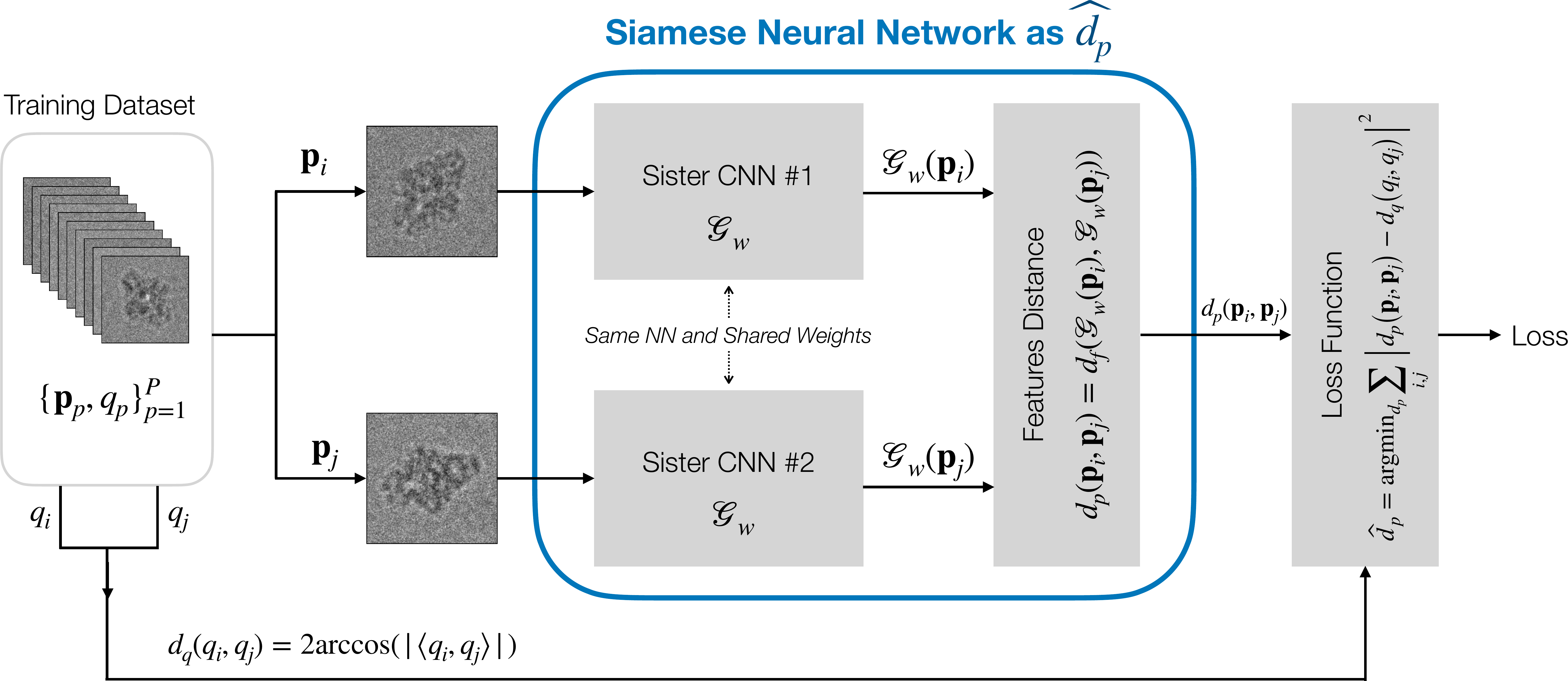}
    \caption{%
        Distance learning.
        We are looking for a distance $\widehat{d_p}$ between projections that is an accurate estimator of the distance $d_q$ between their orientations.
        We propose to parameterize $\widehat{d_p}$ as a Siamese neural network (SiameseNN), trained on a synthetic dataset of projections with associated orientation.
}\label{fig:schematic:distance-learning}
\end{figure}

From a training dataset ${\{ \p_i, q_i \}}_{i=1}^{P}$, we learn the projection distance
\begin{equation}
    \widehat{d_p} = \argmin_{d_p} L_\text{DE},
    \quad \text{where} \quad
    L_\text{DE} = \sum_{i,j} \left| d_p\big(\p_i,\p_j\big) - d_q\big(q_i,q_j\big) \right|^2
    \label{eqn:distance-learning}
\end{equation}
is the loss and $d_q$ is defined in~\eqnref{distance:orientations}.
The distance $d_p$ is parameterized as the Siamese neural network (SiameseNN)~\cite{chopra2005learning}
\begin{equation*}
    d_p(\p_i, \p_j) = d_f(\G_w(\p_i), \G_w(\p_j)),
\end{equation*}
where $\G_w$ is a convolutional neural network with weights $w$ that is trained to extract the most relevant features $\f_i \in \R^{n_f}$ from a projection $\p_i$.
SiameseNNs, also termed ``twin networks'', are commonly used in the field of deep metric learning to learn similarity functions~\cite{yi2014deep}.
We set the feature space distance $d_f$ as the cosine distance to facilitate the learning of a $\widehat{d_p}$ that respects the elliptic geometry of $\mathbb{S}^3$ (\apxref{siamese:feature-distance-and-embedding-dimension}).
\figref{schematic:distance-learning} illustrates the proposed learning paradigm.

As evaluating a sum over $P^2$ pairs is computationally intractable for cryo-EM datasets with typically $P$ in the order of $10^5$ projections, we sample the sum and minimize \eqnref{distance-learning} with stochastic gradient descent (SGD) over small batches of pairs.
The weights $w$ are updated by back-propagation.

The architecture of $\G_w$ is described in \apxref{siamese-architecture}.
When designing the architecture, we constrain the functional space from which the trained $\G_w$ is drawn and express our prior expert knowledge.
For example, we realize shift invariance, \ie, a guarantee that a shift $\mathbf{S_t}$ does not change our estimated distances and orientations, with a fully convolutional architecture.
Size invariance, \ie, taking projections $\p$ of varying sizes $n_p$ while yielding a representation $\f$ of a fixed size $n_f$, is realized by a final average pooling layer.
As we do not (yet) know how to realize an invariance to noise or PSF, we resort to data augmentation, \ie, training on perturbed projections.
In \secref{results:distance-estimation:sensitivity}, we show that a built-in invariance (shift) is far preferable to one learned through augmentation (noise).
Finally, as projections are made by integrating through the 3D volume, projections from opposed directions $(\theta_2,\theta_1)$ are mirrors of each other.%
\footnote{That fact prevents cryo-EM reconstruction to resolve chirality, \ie, it cannot distinguish a protein from its mirrored version.}
That is another kind of physical knowledge that should ideally be built into our method.

One could hope to train $\mathcal{G}_w$ to directly map projections to orientations as $\widehat{q_i} = \f_i = \G_w(\p_i)$.
While that would avoid the orientation recovery step, a space of $n_f=4$ dimensions does not have room for $\mathcal{G}_w$ to represent the other factors of variation in $\p$, such as different noise levels, PSFs, or proteins.
We tested that hypothesis in \apxref{siamese:feature-distance-and-embedding-dimension}.

\subsection{Orientation recovery}\label{sec:method:orientation-recovery}

The task of recovering points based on their relative distances has been extensively studied.
Many methods aim at mapping high-dimensional data onto a lower-dimensional space while preserving distances, primarily for dimensionality reduction and data visualization.
Well-known examples include multi-dimensional scaling (MDS)~\cite{cox2008mds}, Isomap~\cite{tenenbaum2000isomap}, locally linear embedding (LLE)~\cite{roweis2000lle}, Laplacian eigenmaps~\cite{belkin2003laplacian}, t-distributed stochastic neighbor embedding (t-SNE)~\cite{maaten2008tsne}, and uniform manifold approximation and projection (UMAP)~\cite{mcinnes2018umap}.
The embedding of distance matrices in Euclidean space (given by their eigenvectors) is especially well-described.
In particular, the framework of Euclidean distance matrices (EDMs)~\cite{dokmanic2015edm} provides theoretical guarantees on the recovery of points from distances.

We however aim to embed the orientations $q$ in $\mathbb{S}^3$ (\secref{method:orientation-representation}), a setting for which we are unaware of any theoretical characterization (\eg, on the shape of the loss function or its behavior when distances are missing or noisy).
The fact that $\mathbb{S}^3$ is locally Euclidean does however offer some hope. %
Indeed, despite the non-convexity and the lack of theoretical guarantees, we are able to appropriately minimize our loss function, as we experimentally demonstrate in \apxref{results:orientation-recovery:exact}.

We recover the orientations of a set of projections $\big\{ \mathbf{p}_k \big\}_{k=1}^P$ through
\begin{equation}
    \big\{ \widehat{q_k} \big\}_{k=1}^P = \argmin_{\{q_k \in \mathbb{S}^3\}} L_\text{OR},
    \quad \text{where} \quad
    L_\text{OR} = \sum_{i,j} \left| \widehat{d_p} \left( \p_i, \p_j \right) - d_q\left(q_i,q_j\right) \right|^2
    \label{eqn:orientation-recovery}
\end{equation}
is the loss and $\widehat{d_p}$ is the estimator trained in \eqnref{distance-learning}.
Note that the sole difference with~\eqnref{distance-learning} is that the minimization is performed over the orientations $q$ rather than the distance $d_p$.
Here again, we sample the sum in practice and minimize \eqnref{orientation-recovery} with mini-batch SGD.
Sampling the sum amounts to building a sparse (instead of complete) distance graph before embedding, a common strategy.

\subsection{Evaluation}\label{sec:method:evaluation}

While not a part of the method \textit{per se}, the evaluation of the orientations recovered by~\eqnref{orientation-recovery} is essential for assessing the quality of the obtained results.
Unfortunately, we cannot directly take the difference between the recovered orientations $\{\widehat{q_k}\}_{k=1}^P$ and the true orientations $\{q_k\}_{k=1}^P$ as orientations are rotations up to an arbitrary reference orientation.
Any global rotation or reflection of the recovered orientations is as valid as any other, \ie, $d_q(q_i, q_j) = d_q(\T q_i , \T q_j) \; \forall \, \T \in \Or(4)$, where $\Or(4)$ is the group of $4 \times 4$ orthogonal matrices. %
Hence, we align the sets of orientations and compute the \textit{mean orientation recovery error} as
\begin{equation}
    E_\text{OR} = \min_{\T \in \Or(4)} \frac{1}{P} \sum_{i=1}^P \big| d_q\left( q_i, \T \widehat{q_i} \right) \big|.
    \label{eqn:orientation-recovery-error}
\end{equation}
We implement $\T$ as a product of $\binom{4}{2}=6$ independent rotations and an optional reflection:
\begin{equation*}
    \T =
    \begin{bmatrix}
        m & \mathbf{0} \\
        \mathbf{0} & \mathbf{I} \\
    \end{bmatrix}
    \prod_{1 \leq i < j \leq 4} \mathbf{T}_{\theta_{ij}},
    \quad m \in \{-1,1\}, \; \theta_{ij} \in [0, 2\pi[,
\end{equation*}
where $m = \det(\T) = -1$ if $\T$ includes a reflection, and $\mathbf{T}_{\theta_{ij}} \in \SO(4)$ is a rotation by angle $\theta_{ij}$ on the $(x_i, x_j)$ plane.

In practice, we again minimize \eqnref{orientation-recovery-error} with mini-batch SGD.
Because $\Or(4)$ is disconnected, we optimize the 6 angles separately for $m = 1$ (proper rotations) and $m = -1$ (improper rotations).
\figref{5j0n-aa-loss-perfect-distances} shows an alignment to $E_{OR}=0$ after a perfect recovery.

\section{Experiments}\label{sec:experiments}

We first evaluated whether orientation recovery through \eqnref{orientation-recovery} was feasible assuming perfect distances, and how it was affected by errors in the distances (\secref{results:orientation-recovery:sensitivity}).
We then learned to estimate the distances through \eqnref{distance-learning} and evaluated the accuracy of this procedure (\secref{results:distance-estimation:learned}).
Following this, we evaluated the robustness of distance learning to perturbations in the projections (\secref{results:distance-estimation:sensitivity}).
Finally, we ran the whole machinery on a synthetic dataset to assess how well orientations could be recovered from distances estimated by the trained SiameseNN (\secref{results:orientation-recovery:reconstruction}).

\subsection{Experimental conditions}\label{sec:results:data}

\begin{figure}[ht!]
    \centering
    \begin{minipage}[b]{0.51\linewidth}
        \centering
        \begin{subfigure}[b]{0.49\linewidth}
            \centering
            \includegraphics[height=3cm]{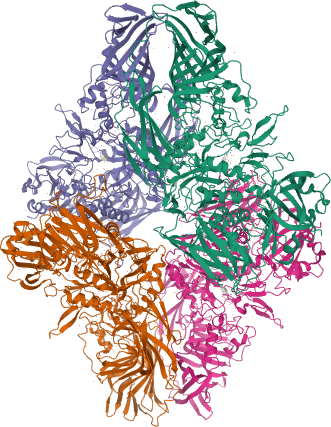}
            \caption{Atomic model of \texttt{5a1a}.}
        \end{subfigure}
        \hfill
        \begin{subfigure}[b]{0.49\linewidth}
            \centering
            \includegraphics[height=3cm]{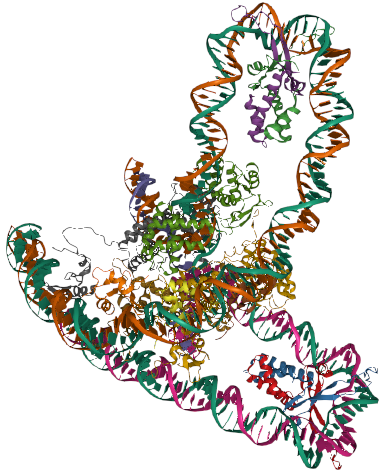}
            \caption{Atomic model of \texttt{5j0n}.}
        \end{subfigure}
        \\ \vspace{0.5em}
        \begin{subfigure}[b]{0.49\linewidth}
            \centering
            \includegraphics[height=3cm]{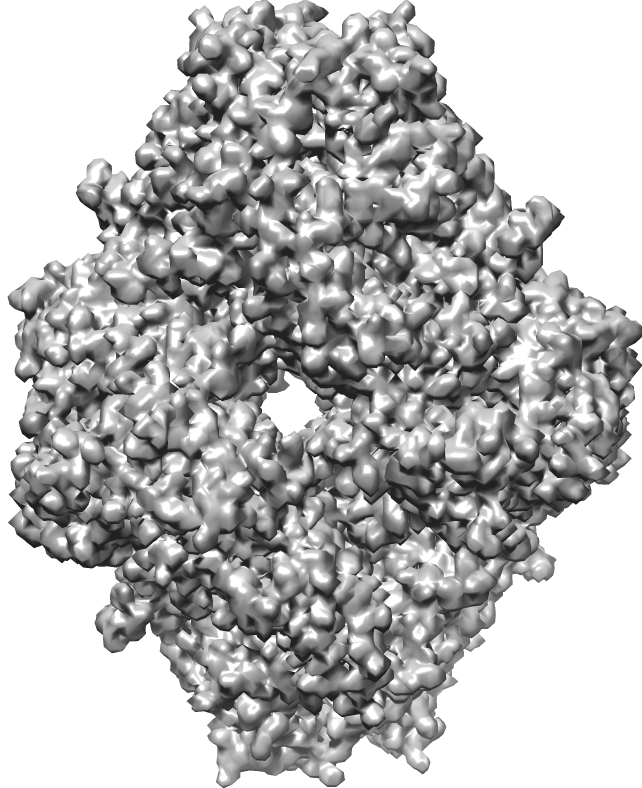}
            \caption{1\AA\ density map $\x$ of \texttt{5a1a}.}%
            \label{fig:density-map:5j0n:ground-truth}
        \end{subfigure}
        \hfill
        \begin{subfigure}[b]{0.49\linewidth}
            \centering
            \includegraphics[height=3cm]{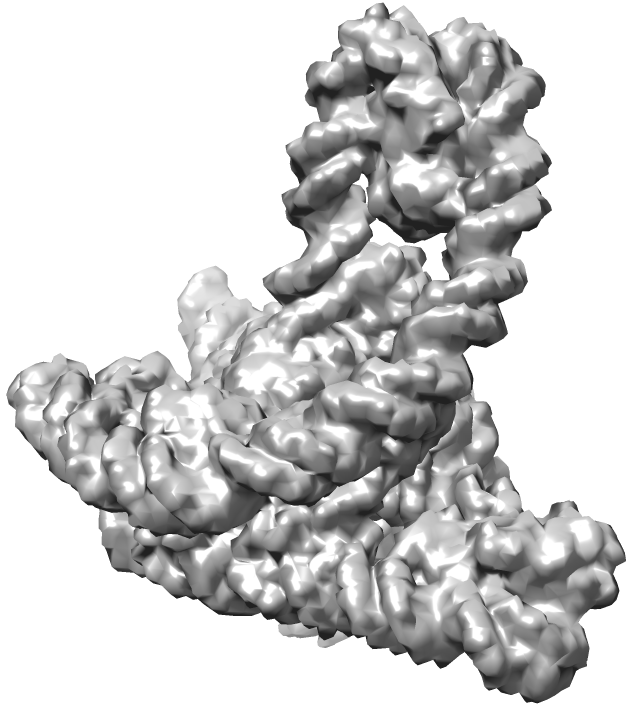}
            \caption{3.67\AA\ density map $\x$ of \texttt{5j0n}.}
        \end{subfigure}
        \caption{%
            Two proteins with different symmetries.
        }\label{fig:pdb-proteins}
    \end{minipage}
    \hfill
    \begin{minipage}[b]{0.48\linewidth}
        \centering
        \begin{subfigure}[b]{0.49\linewidth}
            \centering
            \includegraphics[height=3cm]{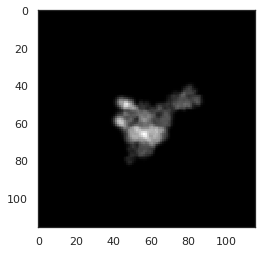}
            \caption{$\p = \mathbf{P}_{\bth} \mathbf{x}$}
        \end{subfigure}
        \hfill
        \begin{subfigure}[b]{0.49\linewidth}
            \centering
            \includegraphics[height=3cm]{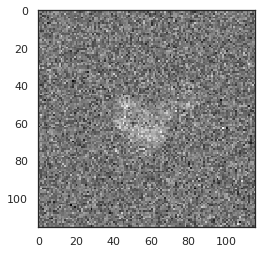}
            \caption{$\p = \mathbf{P}_{\bth} \mathbf{x} + \mathbf{n}$}
        \end{subfigure}
        \\ \vspace{0.5em}
        \begin{subfigure}[b]{0.49\linewidth}
            \centering
            \includegraphics[height=3cm]{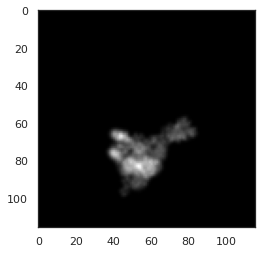}
            \caption{$\p = \mathbf{S}_{\mathbf{t}} \mathbf{P}_{\bth} \mathbf{x}$}
        \end{subfigure}
        \hfill
        \begin{subfigure}[b]{0.49\linewidth}
            \centering
            \includegraphics[height=3cm]{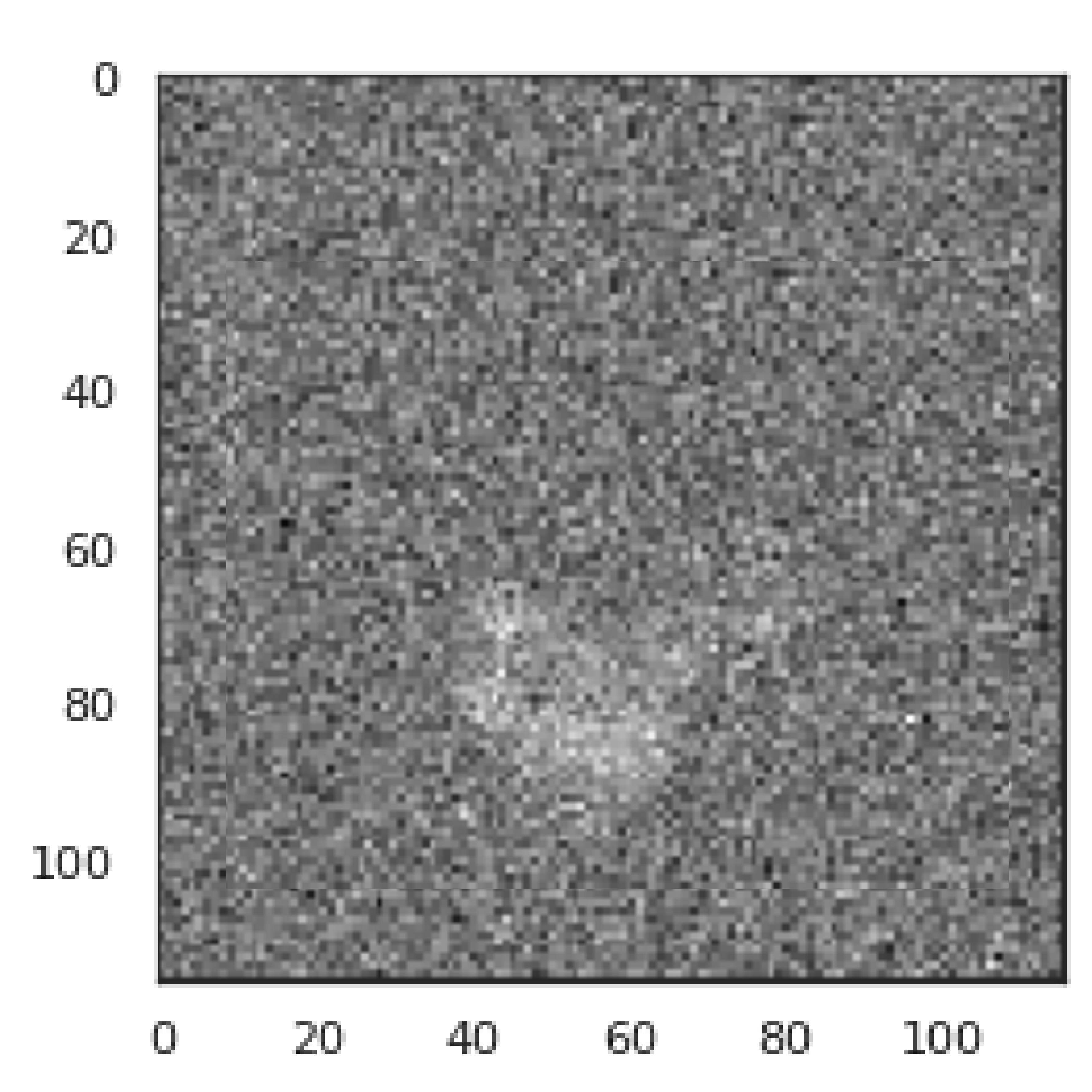}
            \caption{$\p = \mathbf{S}_{\mathbf{t}} \mathbf{P}_{\bth} \mathbf{x} + \mathbf{n}$}
        \end{subfigure}
        \caption{%
            Example projections of \texttt{5j0n} ($\mathbf{n} \sim \mathcal{N}(0, 16\mathbf{I})$).
        }\label{fig:different-projections}
    \end{minipage}
\end{figure}

\paragraph{Density maps.}
We considered two proteins (\figref{pdb-proteins}): the $\beta$-galactosidase, a protein with a dihedral (D2) symmetry, and the lambda excision HJ intermediate (HJI), an asymmetric protein with local cyclic (C1) symmetry.
Their deposited PDB atomic models are \texttt{5a1a}~\cite{bartesaghi2015betagal} and \texttt{5j0n}~\cite{laxmikanthan2016structure}, respectively.
From these atomic models, we generated the density maps in Chimera~\cite{pettersen2004ucsf} by fitting the models with a 1\AA\ map for \texttt{5a1a} and a 3.67\AA\ map for \texttt{5j0n}; this gave us a volume of $110 \times 155 \times 199$ voxels for \texttt{5a1a} and one of $69 \times 57 \times 75$ voxels for \texttt{5j0n}.

\paragraph{Protein symmetries.}
Symmetries are problematic when learning distances: two projections can be identical while not originating from the same orientation, which breaks an axiom of proper distance functions (identity of indiscernibles).
\figref{euclidean-not-robust:5a1a-full} illustrates this problem.
To capture only one of four identical projections of \texttt{5a1a}, we restricted directions to $(\theta_2, \theta_1) \in [0, \pi[ \, \times \, [0, \frac{\pi}{2}[$ (a quarter of the sphere, illustrated in \figref{orientation-sampling:directions}) for that protein.
This treatment of symmetries is incomplete%
\footnote{The remaining issue is that one of four distances is arbitrarily chosen per pair of projections.}
but sufficient for a proof-of-concept.

\paragraph{Projections.}
Using the ASTRA projector~\cite{van2015astra}, we generated $P=5,000$ synthetic projections of $275 \times 275$ pixels (then interpolated to $116 \times 116$) for \texttt{5a1a} and $116 \times 116$ pixels for \texttt{5j0n}, taken from uniformly sampled orientations.%
\footnote{Orientations used in \secref{results:orientation-recovery:sensitivity} (\figref{perfect-with-noise-ar-aa}) and \secref{results:distance-estimation:sensitivity} (\figref{results:distance-estimation}) were actually obtained by uniformly sampling the Euler angles $\bth$, constrained to $(\theta_3,\theta_2,\theta_1) \in [0, 2\pi[ \, \times \, [0, \frac{\pi}{2}[ \, \times \, [0, 2\pi[$ for \texttt{5j0n}. Our conclusions would be identical if orientations were uniformly sampled from $\SO(3)$ instead.}
We then perturbed the measurements with different levels of additive Gaussian noise~\cite{sorzano2004normalizing,shigematsu2013noise} and off-centering shifts. %
\figref{different-projections} displays samples of the simulated projections.

\paragraph{Datasets.}
For each protein, we split the projections into training, validation, and test subsets, and created \textit{disjoint} pairs of projections from each (\tabref{dataset}).
The training and validation sets were used to train and evaluate the SiameseNN, while the test set was used to evaluate orientation recovery given a trained SiameseNN\@.
Sampling orientations (mostly) uniformly induces a distribution of distances that is skewed towards larger distances (shown in \figref{orientation-sampling:distances}).
As this would skew $L_\text{DE}$ and bias $\widehat{d_p}$, we further sampled $1\%$ of the training and validation pairs to make the distribution of distances uniform---for $\widehat{d_p}$ to be uniformly accurate over the whole $[0,\pi]$ range of distances
(see \apxref{orientation-sampling} for further illustrations).
While $1,650$ projections were enough to perfectly reconstruct the density maps (as shown in Figures~\ref{fig:5j0n-reconstruction-fsc} and~\ref{fig:5a1a-reconstruction-fsc}), our method is not limited by the number of projections as optimization is done per batch.

\paragraph{Optimization.}
We optimized \eqnref{distance-learning} with the RMSProp optimizer~\cite{tieleman2012rmsprop} and a learning rate of $10^{-3}$ for $150$ epochs.
Batches of $256$ pairs resulted in $247$ steps per epoch for the training sets and $28$ for the validation sets (\tabref{dataset}).
It took about $3.3$ hours of a single Tesla T4 or $8.75$ hours of a single Tesla K40c.
Our code supports training on multiple GPUs.
We optimized \eqnref{orientation-recovery} with the Adam optimizer~\cite{kingma2014adam} and a learning rate of $0.5$ until convergence on batches of $256$ pairs sampled from the test sets (\tabref{dataset}).
It took about $3.75$ hours of a single Tesla K40c (without early stopping).
We optimized \eqnref{orientation-recovery-error} with the FTRL optimizer~\cite{mcmahan2013ftrl}, a learning rate of $2$, and a learning rate power of $-2$ on batches of $256$ orientations sampled from the test sets (\tabref{dataset}).
We reported the lowest of 6 runs (3 per value of $m$) of 300 steps each.
This took about 50 minutes of CPU\@.
\subsection{Sensitivity of orientation recovery to errors in distance estimation}\label{sec:results:orientation-recovery:sensitivity}

We first evaluated the feasibility of orientation recovery assuming that the exact distances were known.
Experiments confirmed that the method successfully recovers the orientation of every projection in this case (see \apxref{results:orientation-recovery:exact}).

\begin{figure}
    \begin{minipage}[b]{0.48\linewidth}
        \begin{minipage}{\linewidth}
            \begin{tabular}{lrrr}
                \toprule
                Dataset & $P$ & $P^2$ & Used pairs \\
                \midrule
                Training & 2,512 (50\%) & 6,310,144 & 63,101 \\ %
                Validation & 838 (17\%) & 702,244 & 7,022 \\ %
                Test & 1,650 (33\%) & 2,722,500 & 2,722,500 \\ %
                \bottomrule
            \end{tabular}
            \captionof{table}{%
                Split of $P=5,000$ projections in training, validation, and test subsets.
            \vspace{2.5em}
            }\label{tab:dataset}
        \end{minipage}
        \begin{subfigure}[b]{0.49\linewidth}
            \includegraphics[width=\linewidth]{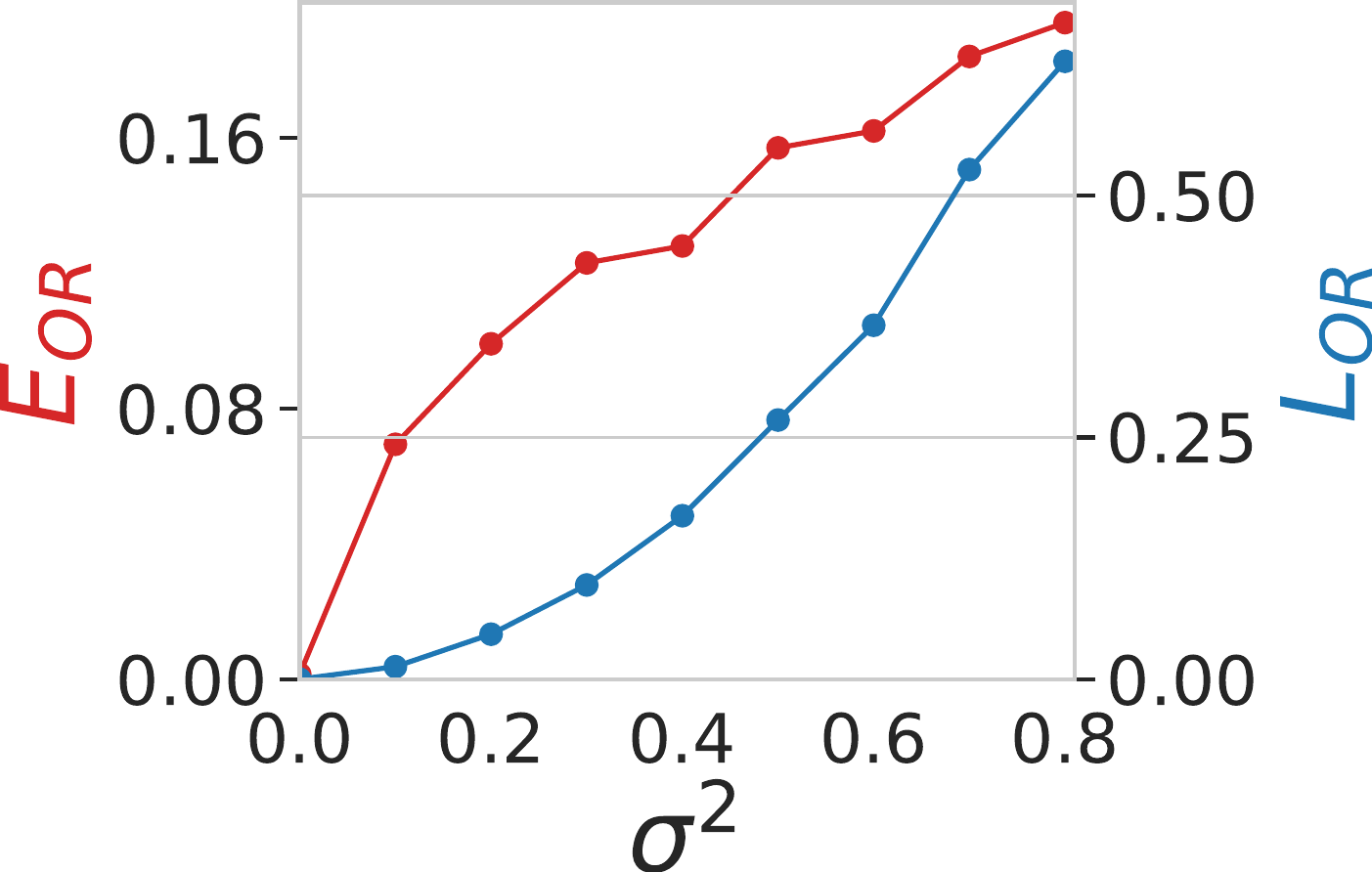}
            \caption{Recovering from \texttt{5j0n}.}
        \end{subfigure}
        \hfill
        \begin{subfigure}[b]{0.49\linewidth}
        \centering
            \includegraphics[width=\linewidth]{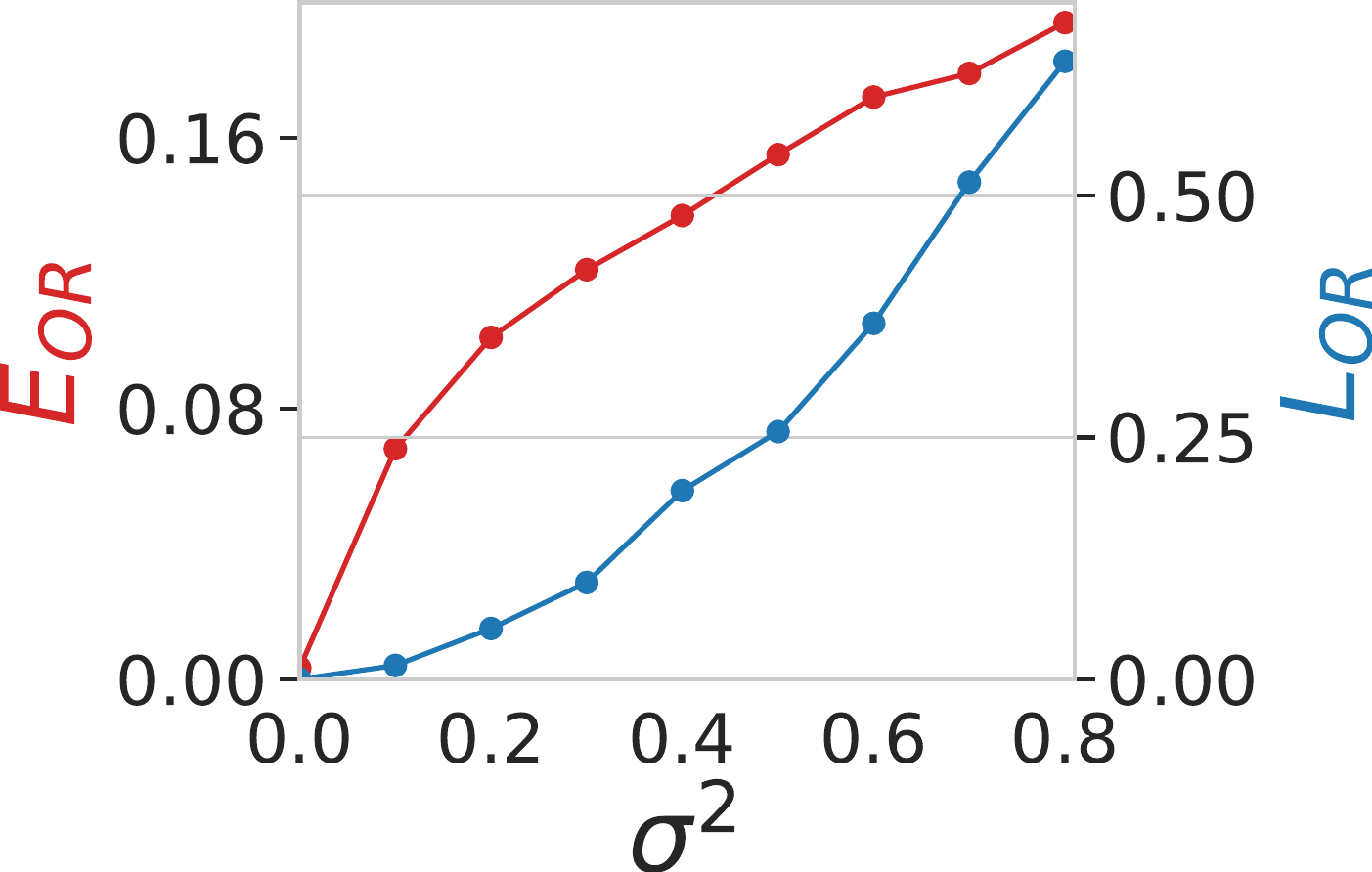}
            \caption{Recovering from \texttt{5a1a}.}
        \end{subfigure}
        \caption{%
            Orientation recovery from perturbed distances.
        }\label{fig:perfect-with-noise-ar-aa}
    \end{minipage}
    \hfill
    \begin{minipage}[b]{0.45\linewidth}
        \begin{subfigure}[b]{\linewidth}
            \centering
            \includegraphics[width=0.47\linewidth]{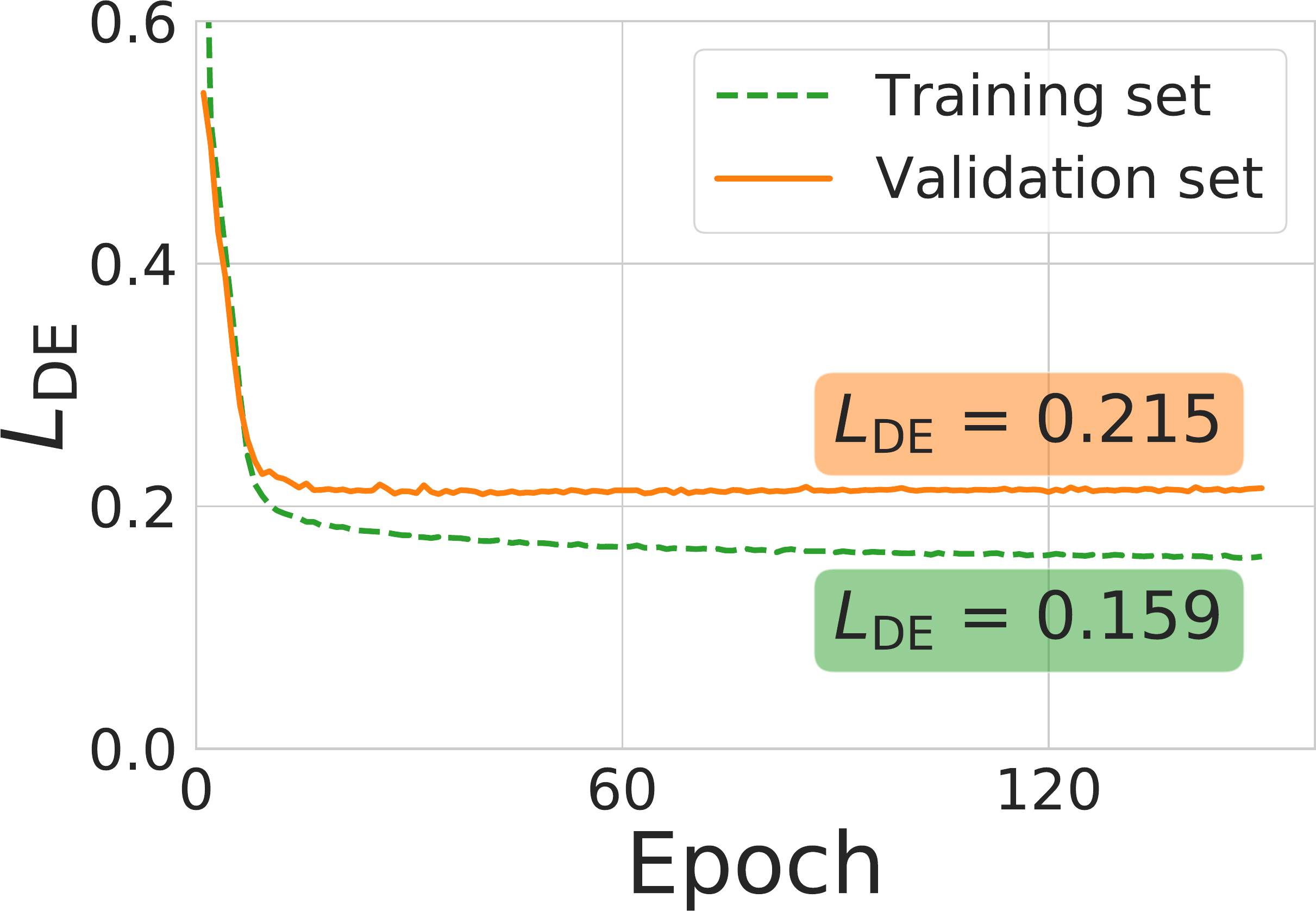}
            \hfill
            \includegraphics[width=0.47\linewidth]{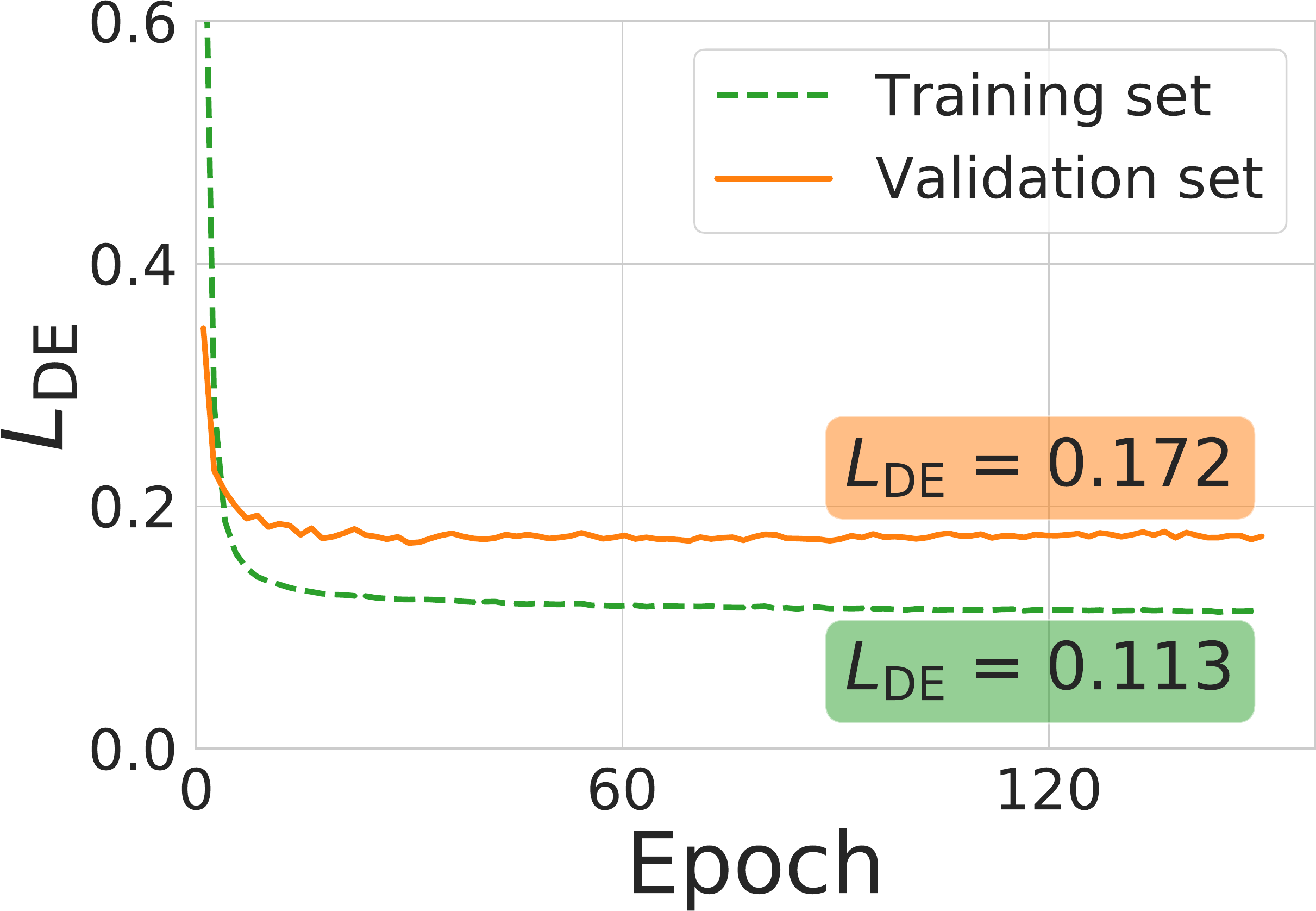}  %
            \caption{Loss converged on \texttt{5j0n} (left) and \texttt{5a1a} (right).
            \vspace{0.8em}}%
            \label{fig:distance-learning:loss}
        \end{subfigure}
        \\ %
        \begin{subfigure}[b]{\linewidth}
            \centering
            \includegraphics[width=0.40\linewidth]{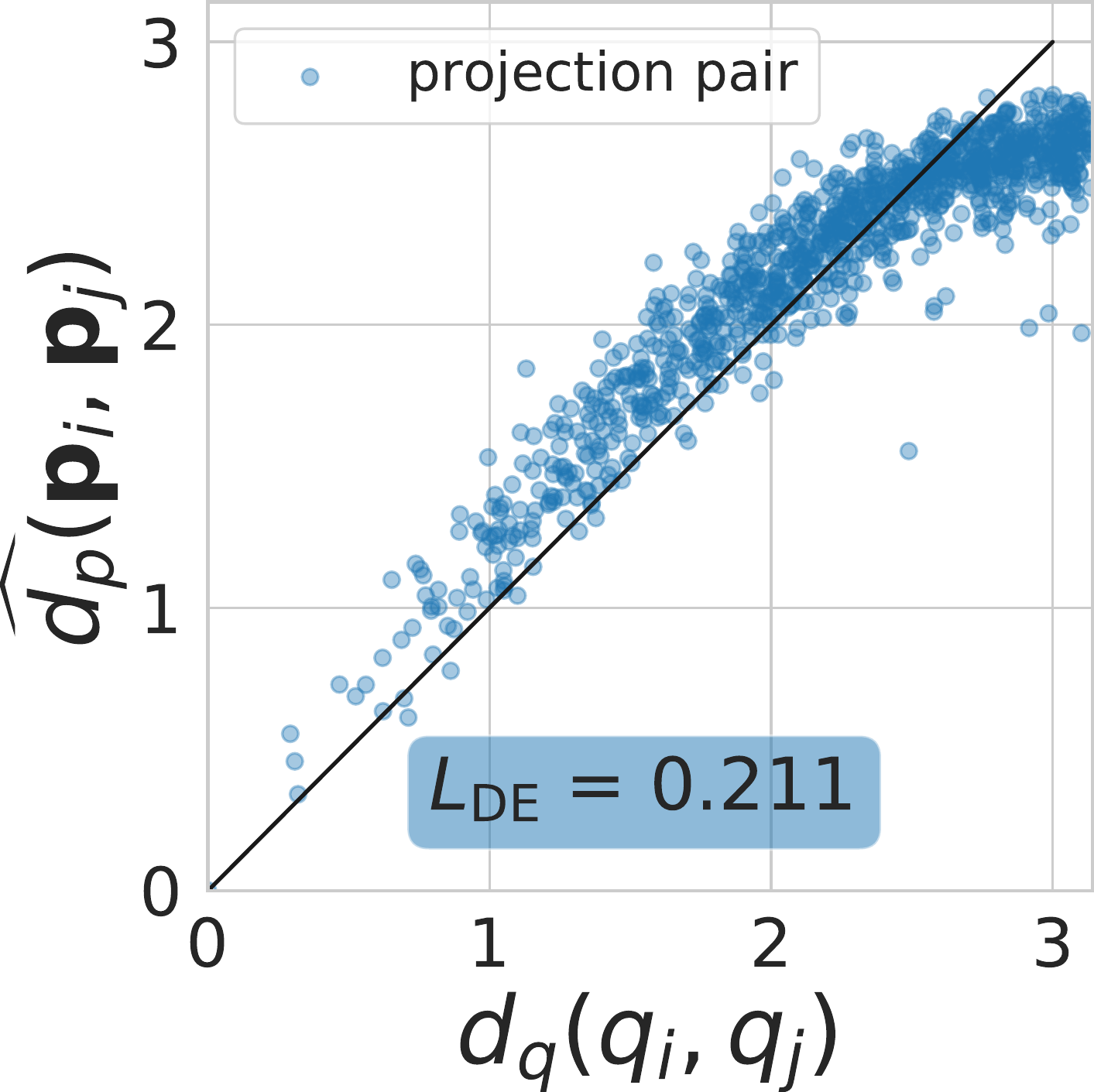}
            \hspace{0.5cm}
            \includegraphics[width=0.40\linewidth]{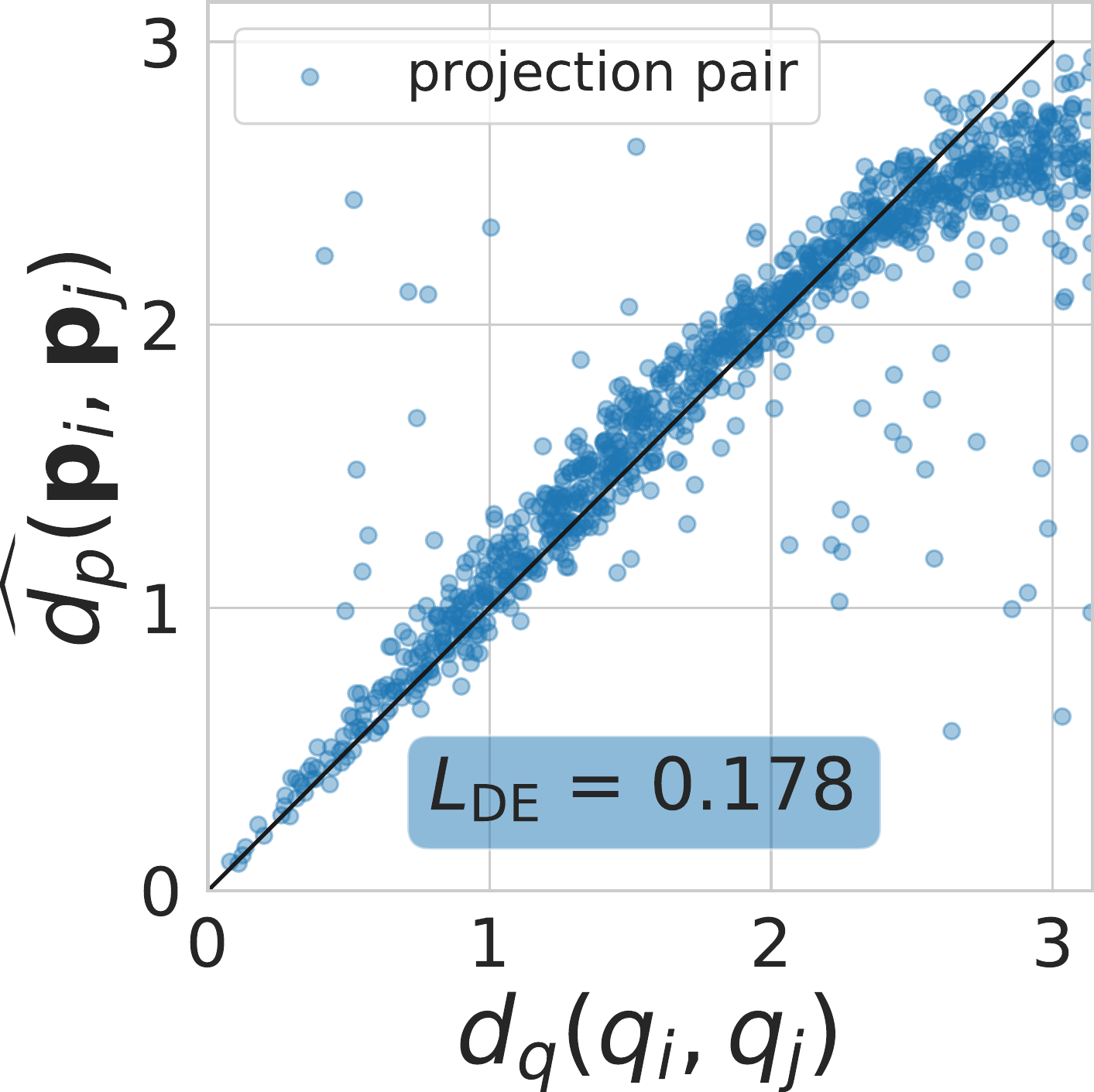}
            \caption{Relationship between $\widehat{d_p}$ and $d_q$ on $1,000$ pairs sampled from the test sets of \texttt{5j0n} (left) and \texttt{5a1a} (right).
            }\label{fig:distance-learning:dpdq}
        \end{subfigure}
        \caption{Distance learning.}
    \end{minipage}
\end{figure}

We then evaluated the behavior of~\eqnref{orientation-recovery} when the true distances were increasingly perturbed.
More precisely, we perturbed the distances prior to the minimization with an error sampled from a Gaussian distribution with mean $0$ and variances $\sigma^2 \in [0.0, 0.8]$.
\figref{perfect-with-noise-ar-aa} shows that the recovery error $E_\text{OR}$ from \eqnref{orientation-recovery-error} is a monotonic function of the error in distances: from $E_\text{OR} = 0$ with perfect distances to $E_\text{OR} \approx 0.2$ radians ($\approx 11.5\degree$) for $\sigma^2 = 0.8$.
These results demonstrate that the performance of orientation recovery~\eqnref{orientation-recovery} depends on the quality of the estimated distances, which advocates for a proper and extensive training of the SiameseNN in further stages of development.

Moreover, we observe that the loss $L_\text{OR}$ from \eqnref{orientation-recovery} is a reliable proxy for $E_\text{OR}$, allowing us to assess recovery performance in the absence of ground-truth orientations (\ie, when recovering the orientations of real projections).

\subsection{Learning to estimate distances}\label{sec:results:distance-estimation:learned}

We evaluated the ability of the SiameseNN to learn to approximate the orientation distance $d_q$ through \eqnref{distance-learning}.
For comparison, we evaluated a baseline, the Euclidean distance $\widehat{d_p}(\p_i,\p_j) = \|\p_i, \p_j\|_2$, in \apxref{results:distance-estimation:euclidean}.

\figref{distance-learning:loss} shows the convergence of $L_\text{DE}$, reached in about 50 epochs.
\figref{distance-learning:dpdq} shows the relationship between the distance $\widehat{d_p}$ estimated from projections and the true distance $d_q$.
The outliers for \texttt{5a1a} are explained by our incomplete treatment of its symmetry.
While our learned distance function is a much better estimator than the Euclidean distance---compare \figref{distance-learning:dpdq} with \figref{euclidean-not-robust}---they share one characteristic: both plateau and underestimate the largest distances.
We did attenuate the phenomenon by sampling training distances uniformly (see \secref{results:data}), and the issue is much less severe than with the Euclidean distance.
An alternative could be to only rely on smaller distances for recovery.
That would however require the addition of a spreading term in \eqnref{orientation-recovery} to prevent the recovered orientations to collapse.

These results confirm that a SiameseNN is able to estimate differences in orientations from projections alone, even though much has yet to be gained from improving upon the rather primitive SiameseNN architecture we are currently using.
The use of additional training data should help further diminish overfitting.

\subsection{Sensitivity of distance learning to perturbations in the projections}\label{sec:results:distance-estimation:sensitivity}
\begin{figure}[ht!]
    \centering
    \begin{subfigure}[t]{0.47\linewidth}
        \includegraphics[width=\linewidth]{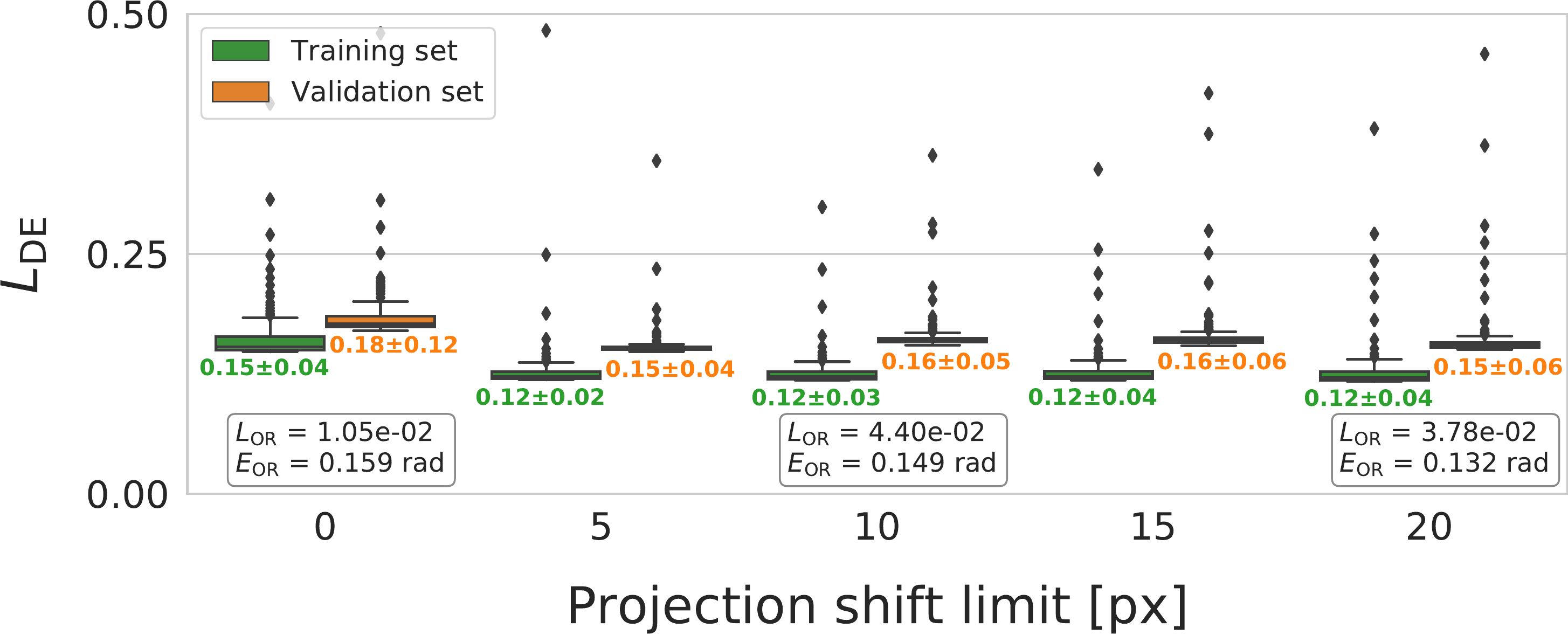}
        \caption{%
            Learning from shifted projections $\{ \mathbf{S}_{\mathbf{t}_i} \mathbf{P}_{\bth_i} \mathbf{x} \}$, with shifts $t_{i_1}$ and $t_{i_2}$ sampled from a triangular distribution with mean 0 and of increasing limits.
            Learning is not harder as projections get shifted farther, because shift invariance is built into the convolutional architecture of $\mathcal{G}_w$.
    }\label{fig:results:distance-estimation:shift}
    \end{subfigure}
    \hfill
    \begin{subfigure}[t]{0.47\linewidth}
        \includegraphics[width=\linewidth]{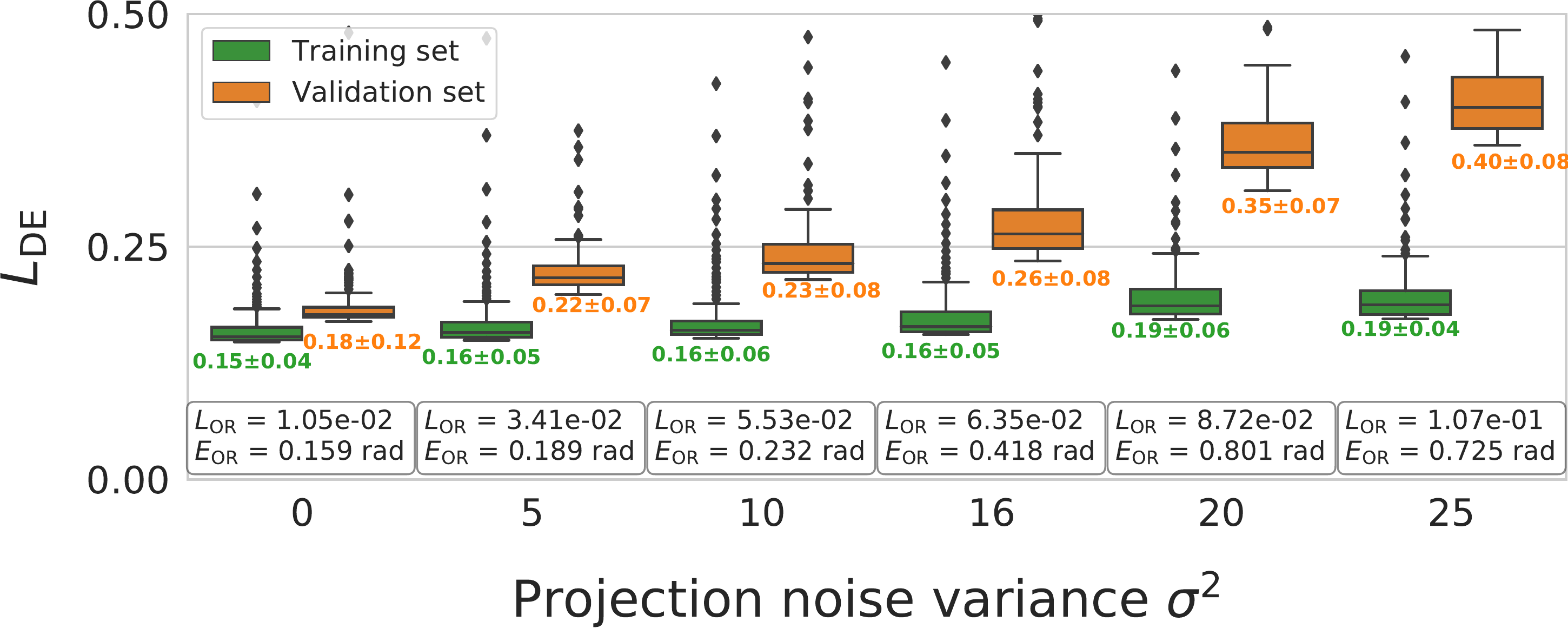}
        \caption{%
            Learning from noisy projections $\{ \mathbf{P}_{\bth_i} \mathbf{x} + \mathbf{n} \}$, with white noise $\mathbf{n} \sim \mathcal{N}(0, \sigma^2\mathbf{I})$ of increasing variance $\sigma^2$.
            Learning is harder as projections get noisier, because noise invariance is not built into the architecture of $\mathcal{G}_w$.
        }\label{fig:results:distance-estimation:noise}
    \end{subfigure}
    \caption{%
        Sensitivity of distance learning to perturbations in the projections of \texttt{5j0n}.
        The box plots show the distance learning loss $L_\text{DE}$~\eqnref{distance-learning} (the distribution is taken over epochs).
        Boxes show the orientation recovery loss $L_\text{OR}$~\eqnref{orientation-recovery} and error $E_\text{OR}$~\eqnref{orientation-recovery-error}.
    }\label{fig:results:distance-estimation}
\end{figure}

We first demonstrated that the learning of distances is insensible to off-centering shifts (\figref{results:distance-estimation:shift}), which is expected given the shift invariance built in our SiameseNN (see \secref{method:distance-learning}).
As we cannot---or do not yet know how to---build noise invariance in the SiameseNN architecture, we trained the SiameseNN on noisy projections and evaluated whether it could learn to treat noise as an irrelevant information.
\figref{results:distance-estimation:noise} shows $E_\text{OR} \approx 0.16$ radians ($\approx 9\degree$) for noiseless projections and $E_\text{OR} \approx 0.42$ radians ($\approx 24\degree$) for a more realistic noise variance of $\sigma^2=16$ (with signal-to-noise ratio of -$12$ dB).
Whereas a naive distance function (\textit{e.g.}, an Euclidean distance) would be extremely sensitive to noise, the SiameseNN mostly learned to discard it.
Moreover, the observed overfitting indicates that more training data should further decrease the sensitivity of the SiameseNN to noise.

Note that we did not evaluate sensitivity to the PSF at this stage but expect a similar behavior.

Here again (\secref{results:orientation-recovery:sensitivity}), we observed that (i) the estimation of more accurate distances (a smaller $L_\text{DE}$) leads to the recovery of more accurate orientations (a smaller $L_\text{OR}$ and $E_\text{OR}$), and that (ii) an higher recovery loss $L_\text{OR}$ induces an higher error $E_\text{OR}$.

\subsection{Orientation recovery and reconstruction of density maps}\label{sec:results:orientation-recovery:reconstruction}

As a proof-of-concept, we attempted to solve the full inverse problem posed by \eqnref{imaging-model}, \ie, to reconstruct the density maps $\widehat{\x}$ from sets of projections $\{ \p_i \}$ and their orientations $\{ \widehat{q_i} \}$ recovered through the proposed method.
It is worth noting that, at this stage of development, we only trained the SiameseNN on projections originating from the protein we were attempting to reconstruct.
In addition, reconstruction was performed with a direct reconstruction algorithm (ASTRA's GPU implementation of the CGLS algorithm) rather than with a robuster iterative method.
This is obviously a specific experimental case that only partially shines light on the applicability of the method in real situations; this is discussed in \secref{discussion}.

\begin{figure}[t]
    \centering
    \begin{subfigure}[b]{0.251\linewidth}
        \centering
        \includegraphics[height=2.4cm]{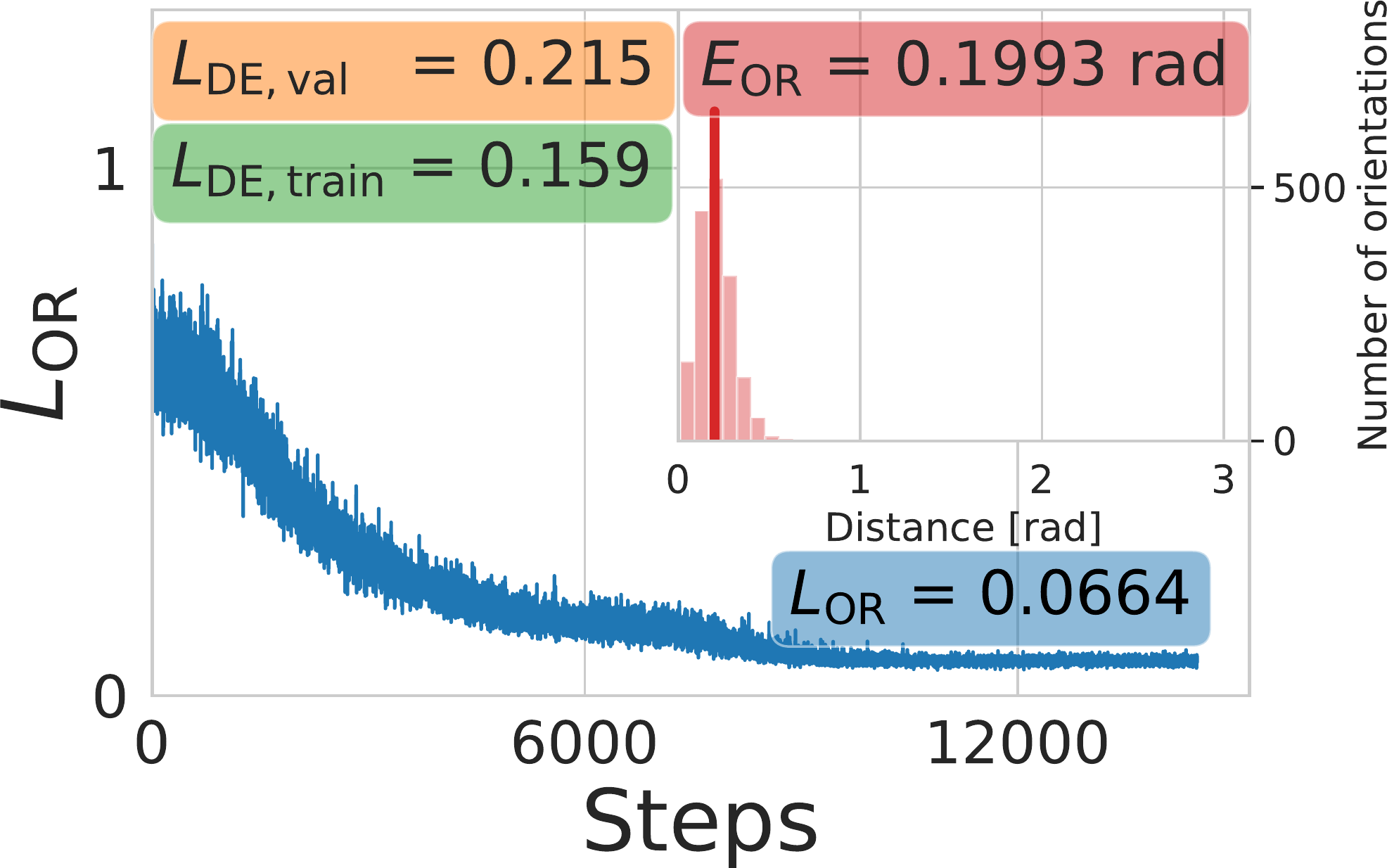}
        \caption{Noiseless projections of \texttt{5j0n}.}%
        \label{fig:5j0n-noise0-orientation-recovery}
    \end{subfigure}
    \hfill
    \begin{subfigure}[b]{0.232\linewidth}
        \centering
        \includegraphics[height=2.4cm]{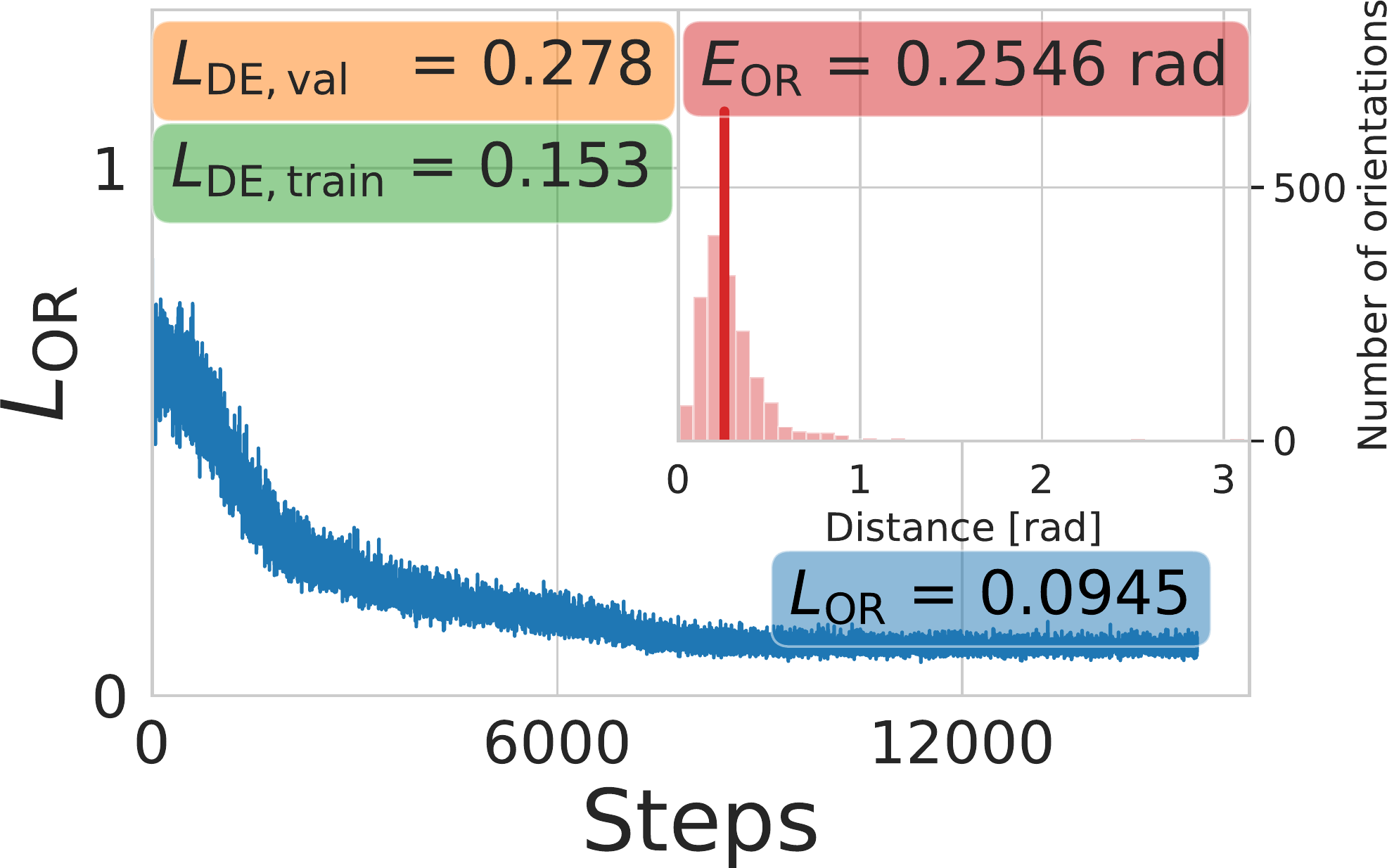}
        \caption{Noisy projections of \texttt{5j0n}.}%
        \label{fig:5j0n-noise16-orientation-recovery}
    \end{subfigure}
    \hfill
    \begin{subfigure}[b]{0.251\linewidth}
        \centering
        \includegraphics[height=2.4cm]{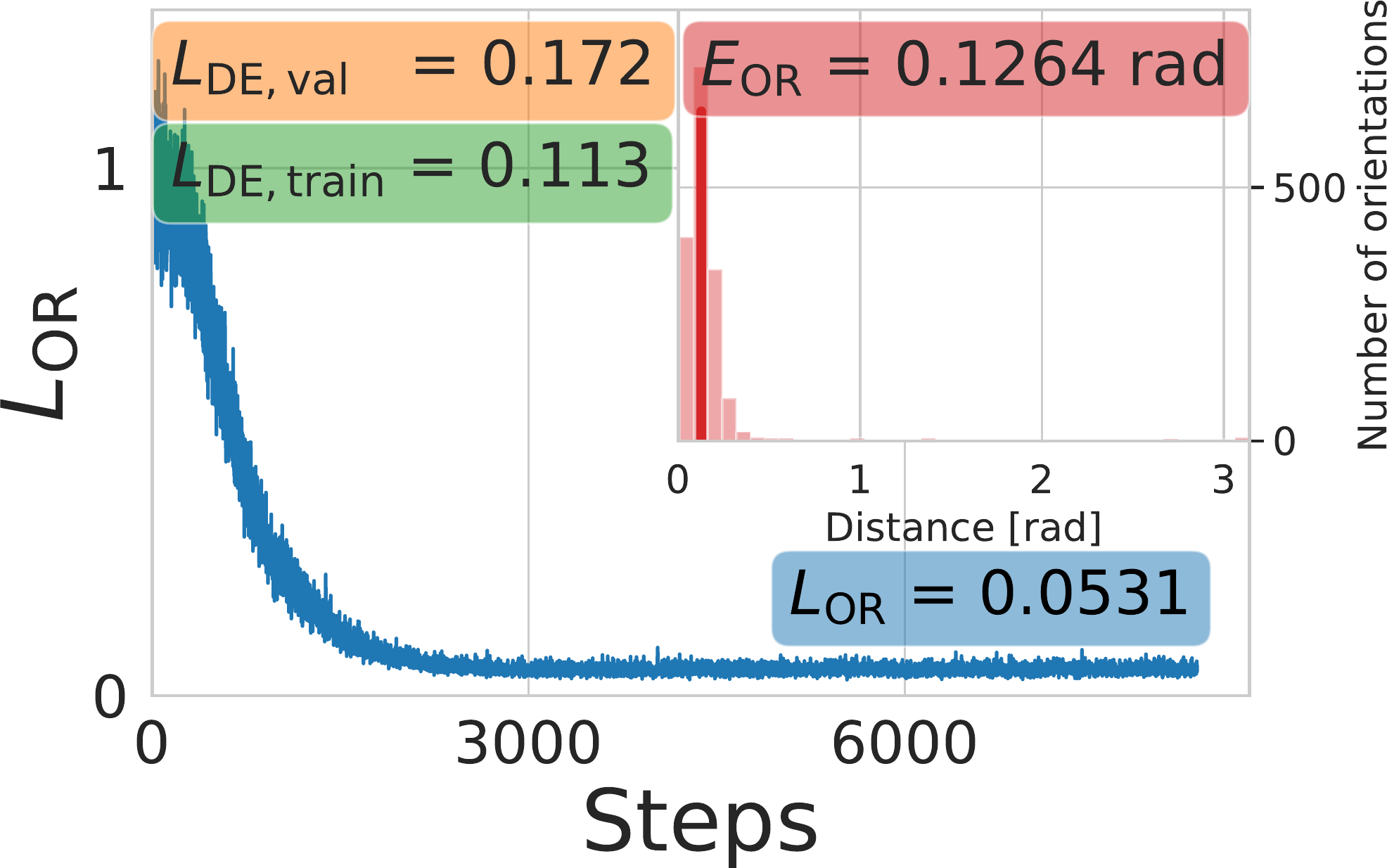}
        \caption{Noiseless projections of \texttt{5a1a}.
        }%
        \label{fig:5a1a-noise0-orientation-recovery}
    \end{subfigure}
    \begin{subfigure}[b]{0.232\linewidth}
        \centering
        \includegraphics[height=2.4cm]{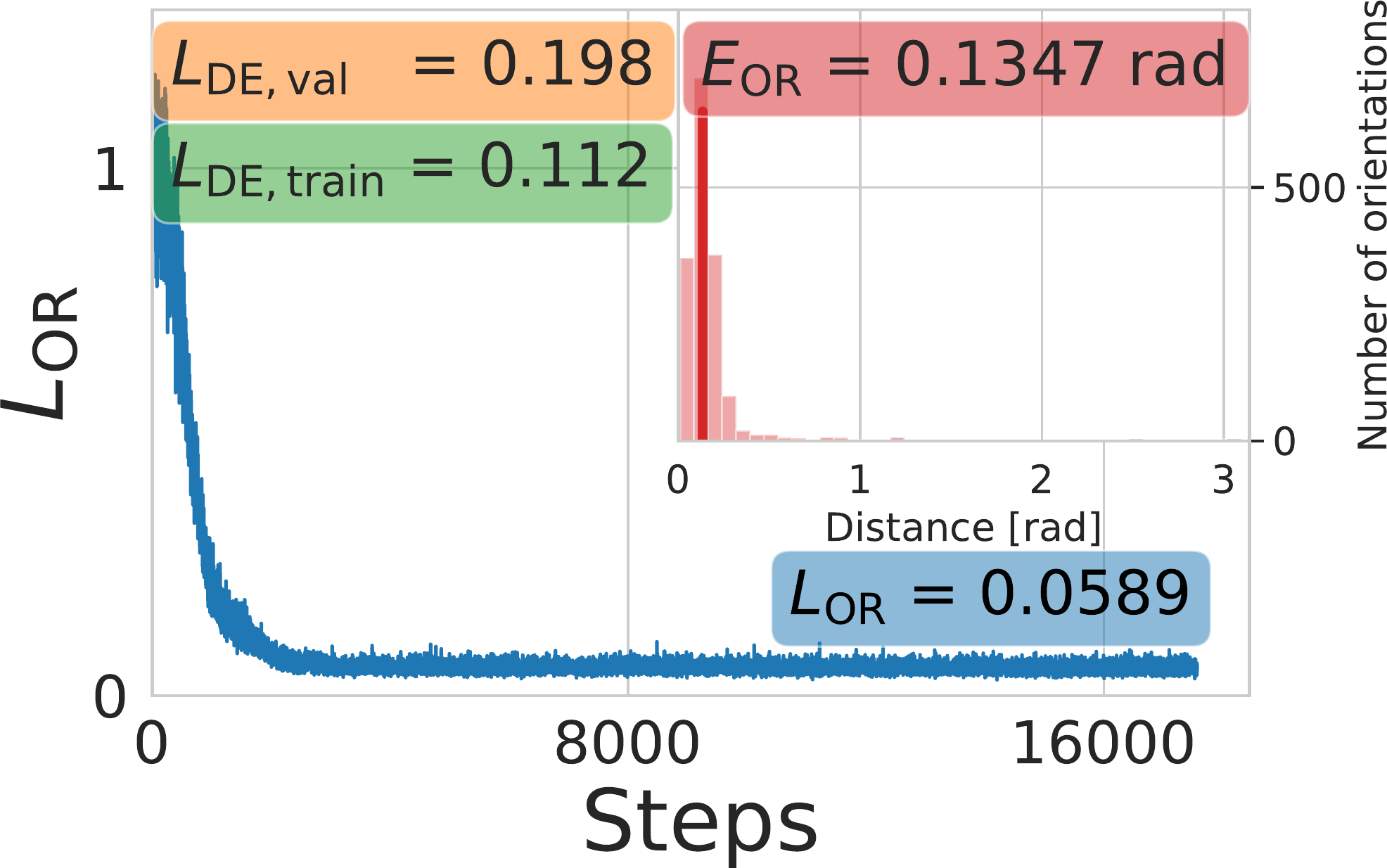}
        \caption{Noisy projections of \texttt{5a1a}.
        }%
        \label{fig:5a1a-noise16-orientation-recovery}
    \end{subfigure}
    \hfill
    \caption{%
        Distance learning and orientation recovery from estimated distances.
        The green and orange boxes show $L_\text{DE}$ \eqnref{distance-learning} on the training and validation sets.
        The blue curve shows the evolution of the recovery loss until convergence, with the minimum $L_\text{OR}$ \eqnref{orientation-recovery} highlighted.
        The red histogram shows the errors in the recovered orientations $\{d_q(q_i, \mathbf{T}\widehat{q_i})\}$, with the mean $E_\text{OR}$ \eqnref{orientation-recovery-error} highlighted.
    }
\end{figure}

\begin{figure}[t]
    \begin{subfigure}[b]{0.17\linewidth}
        \centering
        \includegraphics[height=2.5cm]{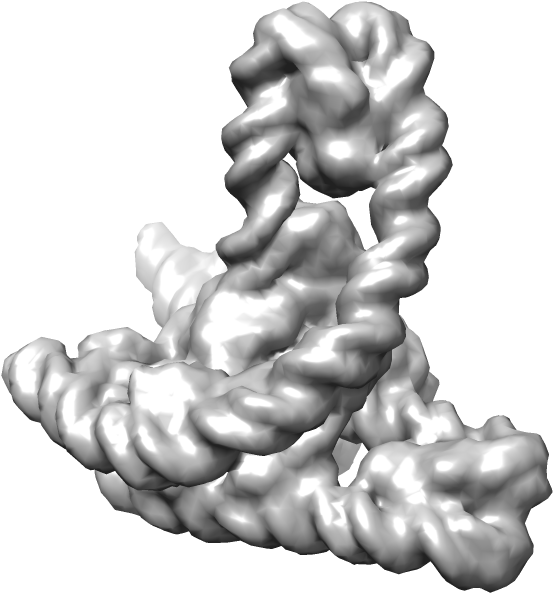}
        \caption{}\label{fig:5j0n-noise0-reconstruction-true}
    \end{subfigure}
    \hfill
    \begin{subfigure}[b]{0.16\linewidth}
        \centering
        \includegraphics[height=2.5cm]{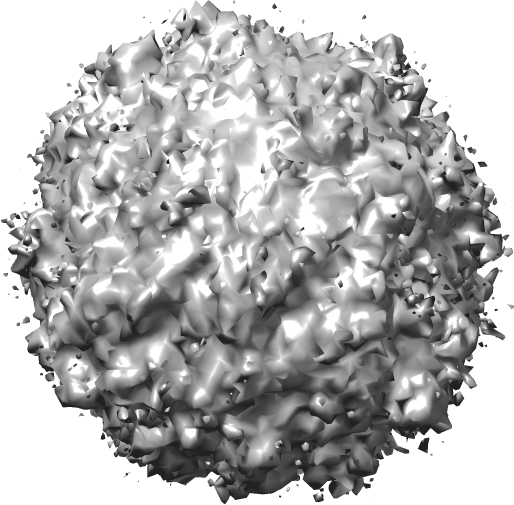}
        \caption{}
    \end{subfigure}
    \hfill
    \begin{subfigure}[b]{0.15\linewidth}
        \centering
        \includegraphics[height=2.5cm]{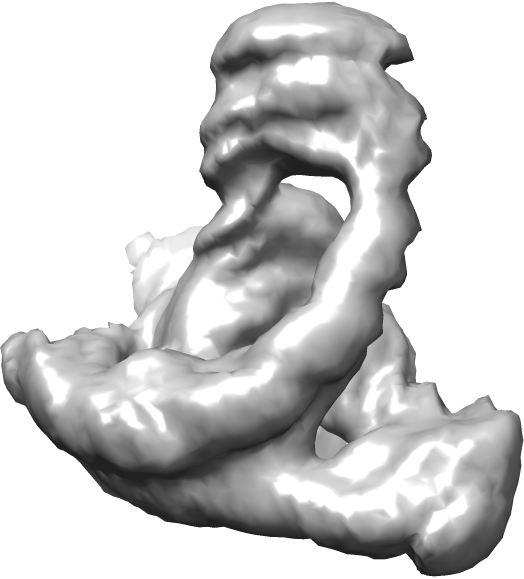}
        \caption{}\label{fig:5j0n-noise0-reconstruction-recovered}
    \end{subfigure}
    \hfill
    \begin{subfigure}[b]{0.15\linewidth}
        \centering
        \includegraphics[height=2.5cm]{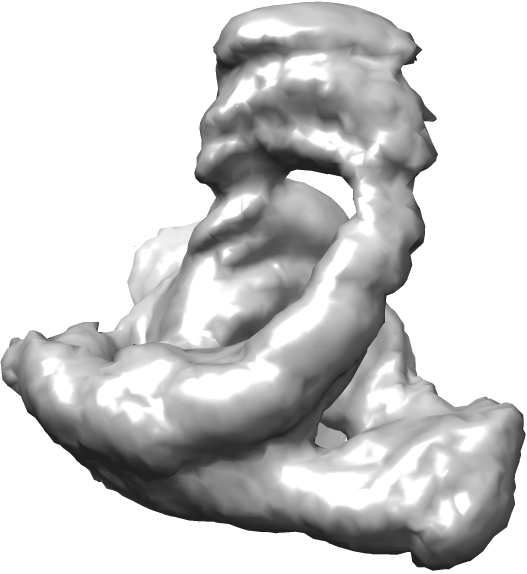}
        \caption{}\label{fig:5j0n-noise16-reconstruction-recovered}
    \end{subfigure}
    \hfill
    \begin{subfigure}[b]{0.30\linewidth}
        \centering
        \includegraphics[height=2.5cm]{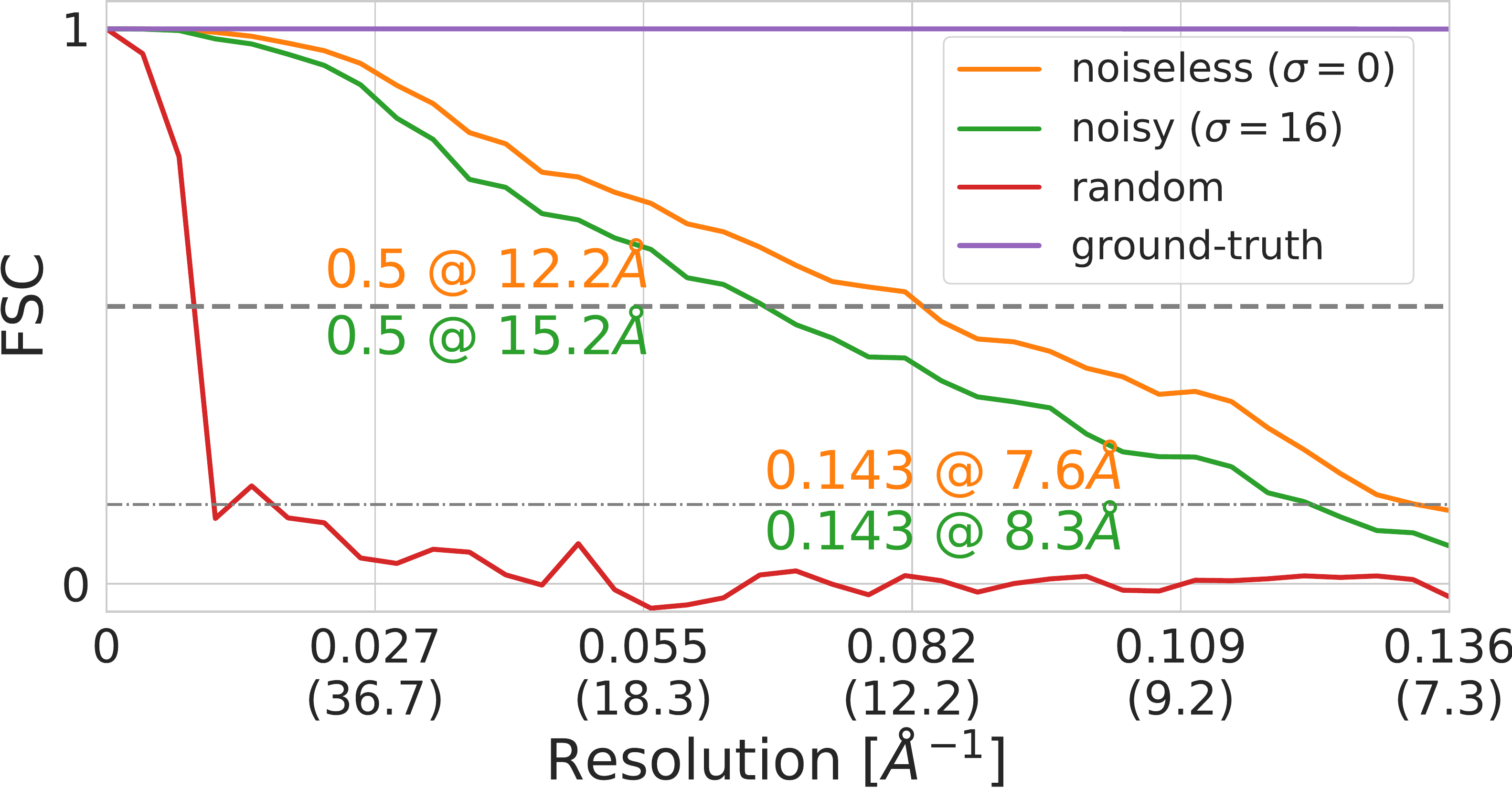}
        \caption{}\label{fig:5j0n-reconstruction-fsc}
    \end{subfigure}

    \vspace{1em}
    \begin{subfigure}[b]{0.17\linewidth}
        \centering
        \includegraphics[height=2.5cm]{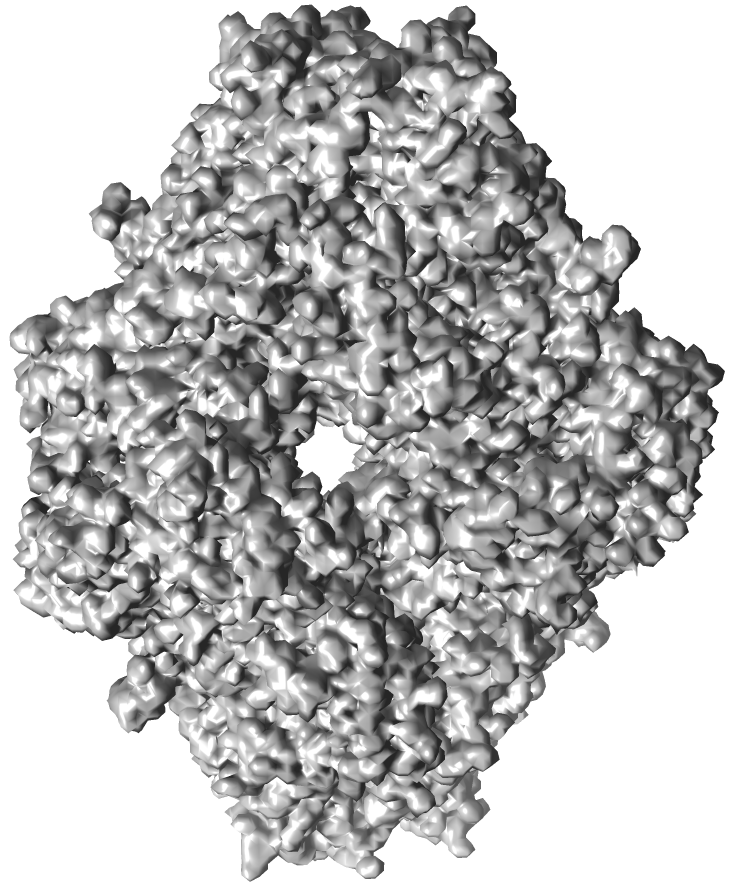}
        \caption{}\label{fig:5a1a-noise0-reconstruction-true}
    \end{subfigure}
    \hfill
    \begin{subfigure}[b]{0.16\linewidth}
        \centering
        \includegraphics[height=2.5cm]{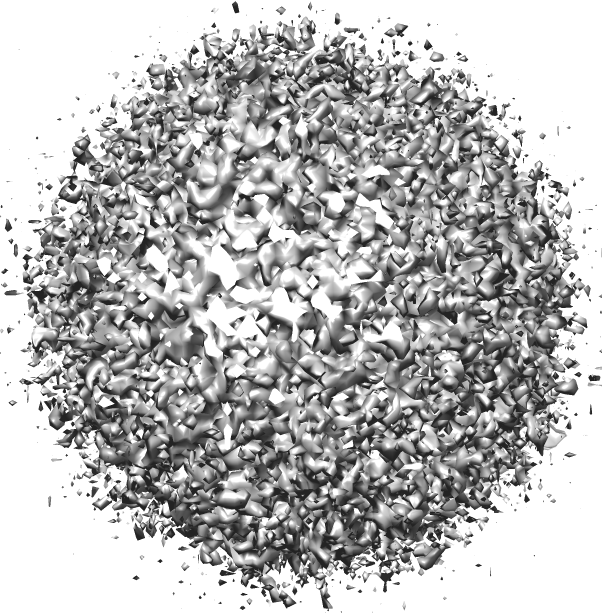}
        \caption{}
    \end{subfigure}
    \hfill
    \begin{subfigure}[b]{0.15\linewidth}
        \centering
        \includegraphics[height=2.5cm]{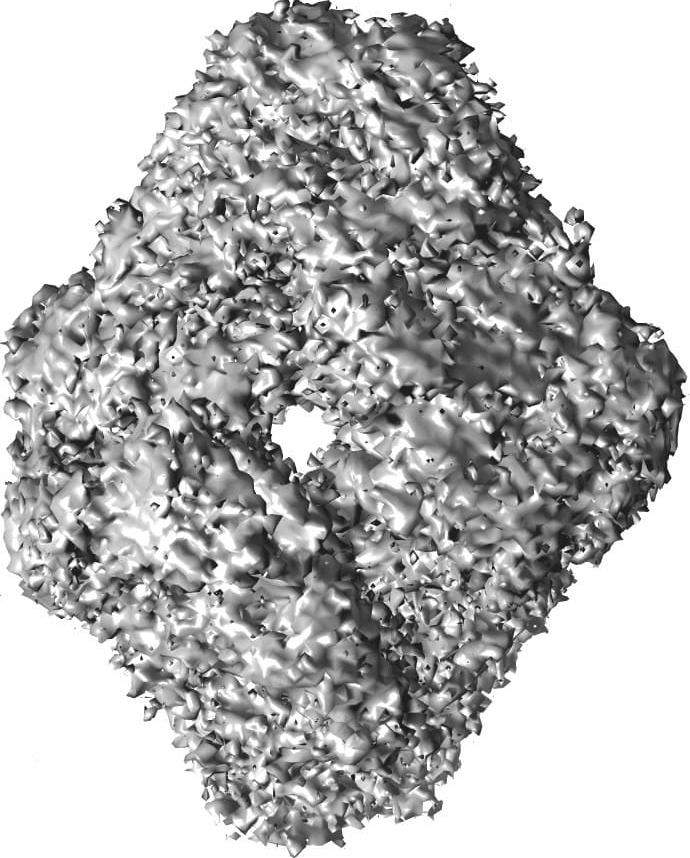}
        \caption{}\label{fig:5a1a-noise0-reconstruction-recovered}
    \end{subfigure}
    \hfill
    \begin{subfigure}[b]{0.15\linewidth}
        \centering
        \includegraphics[height=2.5cm]{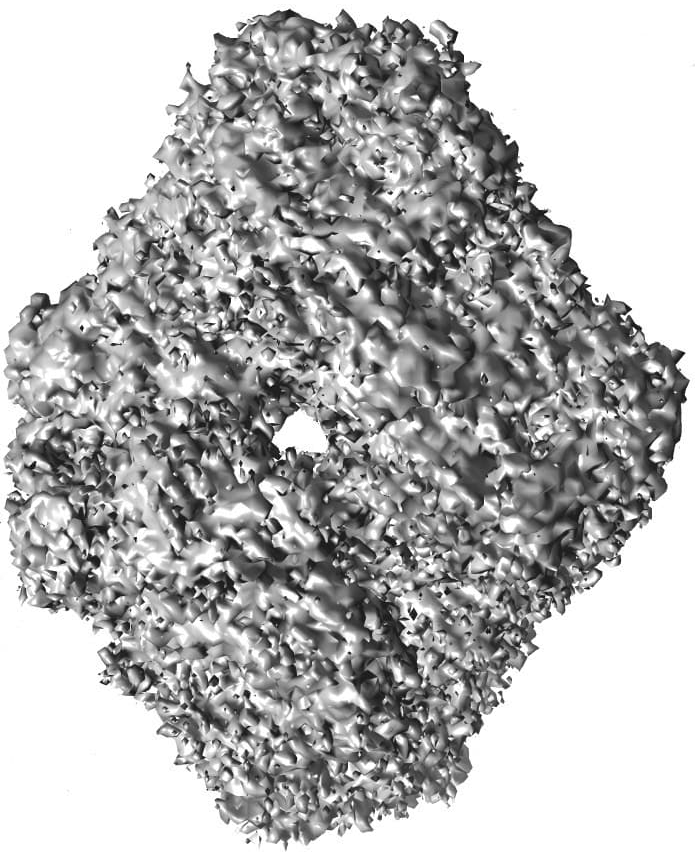}
        \caption{}\label{fig:5a1a-noise16-reconstruction-recovered}
    \end{subfigure}
    \hfill
    \begin{subfigure}[b]{0.30\linewidth}
        \centering
        \includegraphics[height=2.5cm]{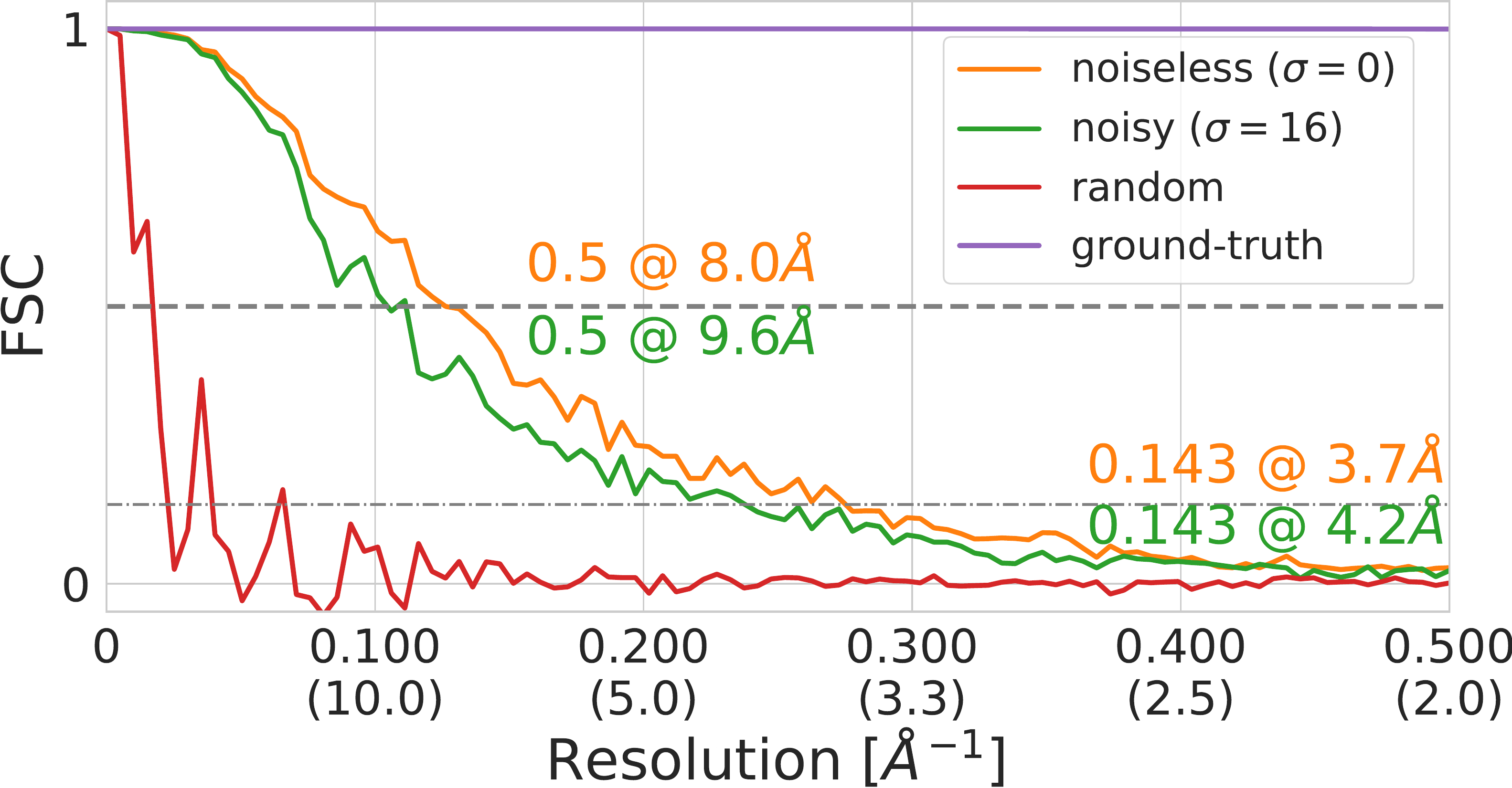}
        \caption{}\label{fig:5a1a-reconstruction-fsc}
    \end{subfigure}
    \caption{%
        Density maps $\widehat{\x}$ reconstructed from (a,f) ground-truth orientations, (b,g) random orientations, (c,h) orientations recovered from noiseless projections, and (d,i) orientations recovered from noisy projections.
        The Fourier shell correlation (FSC) curves in (e,j) indicate the resolutions of the densities reconstructed from recovered orientations (w.r.t.\ ground-truth densities, shown in \figref{density-map:5j0n:ground-truth},d).
    }\label{fig:reconstructions}
\end{figure}

\figref{5j0n-noise0-orientation-recovery} shows the recovery of orientations from distances that were estimated from noiseless projections of \texttt{5j0n}.
A mean error of $E_\text{OR} \approx 0.20$ radians ($\approx 11\degree$) in the recovered orientations led to a reconstruction with a resolution of 12.2\AA\ at a Fourier shell coefficient (FSC) of $0.5$, shown in \figref{5j0n-noise0-reconstruction-recovered}.

As predicted by our other experiments, corrupting the projections with noise ($\sigma^2=16$) negatively impacts the quality of the recovered orientations (\figref{5j0n-noise16-orientation-recovery}); the obtained mean error is then $E_\text{OR} \approx 0.25$ radians ($\approx 14\degree$).
Unsurprisingly, this leads to a reconstruction with a lower resolution of 15.2\AA, shown in \figref{5j0n-noise16-reconstruction-recovered}.
(Note that reconstruction was here obtained from the noiseless projections, the goal being to evaluate only the impact of orientation mis-estimation.)

Finally, \figref{5a1a-noise0-orientation-recovery} and~\ref{fig:5a1a-noise16-orientation-recovery} show the recovery of orientations from noiseless and noisy projections of \texttt{5a1a}.
A mean error of $E_\text{OR} \approx 0.13$ radians ($\approx 7\degree$) in both cases led to reconstructions with resolutions of 8.0\AA\ and 9.6\AA, shown in \figref{5a1a-noise0-reconstruction-recovered} and~\ref{fig:5a1a-noise16-reconstruction-recovered}.
Distance estimation, orientation recovery, and reconstruction performed better on \texttt{5a1a} than \texttt{5j0n} because its ground-truth density is of higher resolution.

These results tend to indicate that a reasonable first structure can be reconstructed from projections whose orientations have been recovered through our method.

\section{Discussion}\label{sec:discussion}

In this work, we explored the use of distance learning between pairs of 2D cryo-EM projections from a 3D protein structure to infer the unknown orientation at which each projection was imaged from.
Our two-step method relies on the training of a SiameseNN to estimate pairwise distances between unseen projections, followed by the recovery of the orientations from these distances through an appropriate minimization scheme.

At the current stage of development, the method has been evaluated on synthetic datasets for two different proteins.
The results provide key insights on the viability of the proposed scheme.
First, they demonstrate that a SiameseNN can learn a distance function between projections that estimates the difference in their orientation (\secref{results:distance-estimation:learned}) and that is invariant to off-centering shifts and robust to increasing levels of noise (\secref{results:distance-estimation:sensitivity})---an important condition in cryo-EM\@.
Second, they demonstrate that an accurate estimation of distances leads to an accurate recovery of orientations (\secref{results:orientation-recovery:sensitivity}, \secref{results:distance-estimation:sensitivity}).
Finally, our method was able to recover orientations with an error of $0.12$ to $0.25$ radians ($7$ to $14\degree$)---leading to an initial volume
with a resolution of $8$ to $15$\AA\ (\secref{results:orientation-recovery:reconstruction}).
In summary, the more accurate the estimated distances, the more precise the recovered orientations, and, ultimately, the higher-resolution the reconstructed volume.

While the method is not yet at the stage where it can be deployed in practice, we believe that a series of developments could help it become a more relevant contributor for single-particle cryo-EM reconstruction.%
\footnote{Note that the present project will not be further continued by its authors due to other professional occupations. Hence, we strongly encourage anyone interested to build on these ideas and, hopefully, make it a practical tool.}
As previously discussed, the results underline the importance of learning an accurate distance estimator. %
In this regard, the performance of the SiameseNN could be improved in several ways.
First, the architecture of the SiameseNN's twin convolutional neural networks should be expanded and tuned.
Second, the training of the SiameseNN could be improved, perhaps by providing more supervision by separately predicting the differences in direction $(\theta_2,\theta_1)$ and in-plane angle $\theta_3$.

Importantly, the SiameseNN would be better trained on a more exhaustive and diverse cryo-EM dataset.
Indeed, the success of the SiameseNN as a faithful estimator of relative orientations eventually relies on our capacity to generate a synthetic training dataset whose data distribution is diverse enough to cover that of unseen projection datasets.
Such realistic cryo-EM projections could be generated by relying on a more expressive formulation of the cryo-EM physics and taking advantage of the thousands of atomic models available in the PDB\@.
In particular, a necessary extension will be to include the effects of the PSF when generating training data and evaluate its impact on the SiameseNN\@. %

A final phase of tests before deploying the method on real cryo-EM measurements will be to extensively test the method on ``unseen proteins'', \ie, proteins whose simulated projections have never been seen by the SiameseNN\@.
In this regard, an interesting aspect of our method is that the twin networks within the SiameseNN intrinsically predict the \textit{relationship} between projections, allowing the SiameseNN as a whole to abstract the particular volume.
Learning benefits from the profound structural similarity shared by proteins---after all, they are all derived from the same $21$ building blocks.

\section*{Acknowledgments}

The authors are thankful to Dr.\ Matthieu Simeoni (EPFL) and Dr.\ Julien Fageot (EPFL) for insightful discussions.

\bibliographystyle{IEEEtran}
\bibliography{refs}

\appendix

\section{Sampling of orientations}\label{apx:orientation-sampling}

\begin{figure}[ht!]
    \centering
    \begin{minipage}{.33\linewidth}
        \begin{subfigure}[b]{\linewidth}
            \centering
            \includegraphics[height=2.5cm]{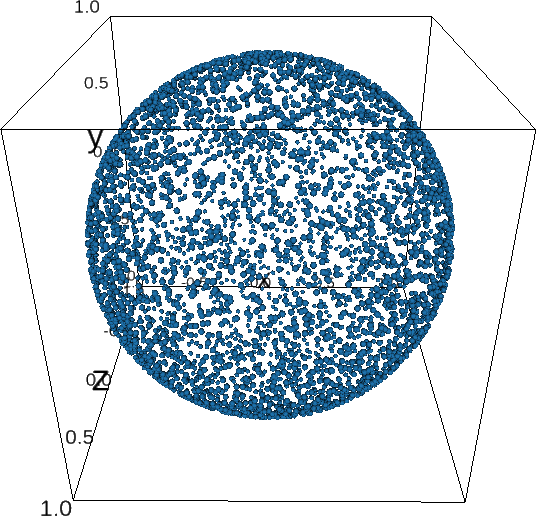}%
            \hfill
            \includegraphics[height=2.5cm]{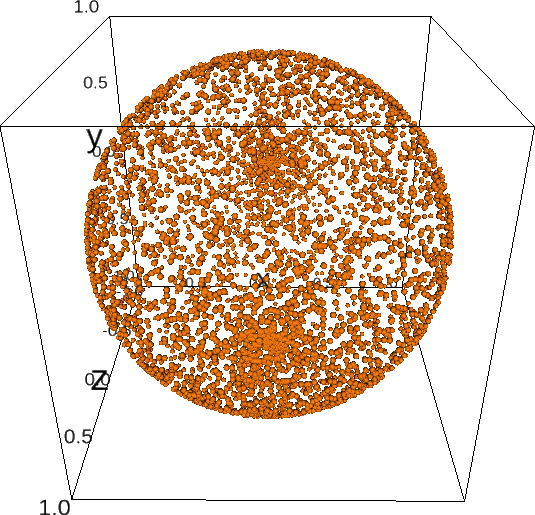}
            \\ \vspace{0.2cm}
            \includegraphics[height=2.5cm]{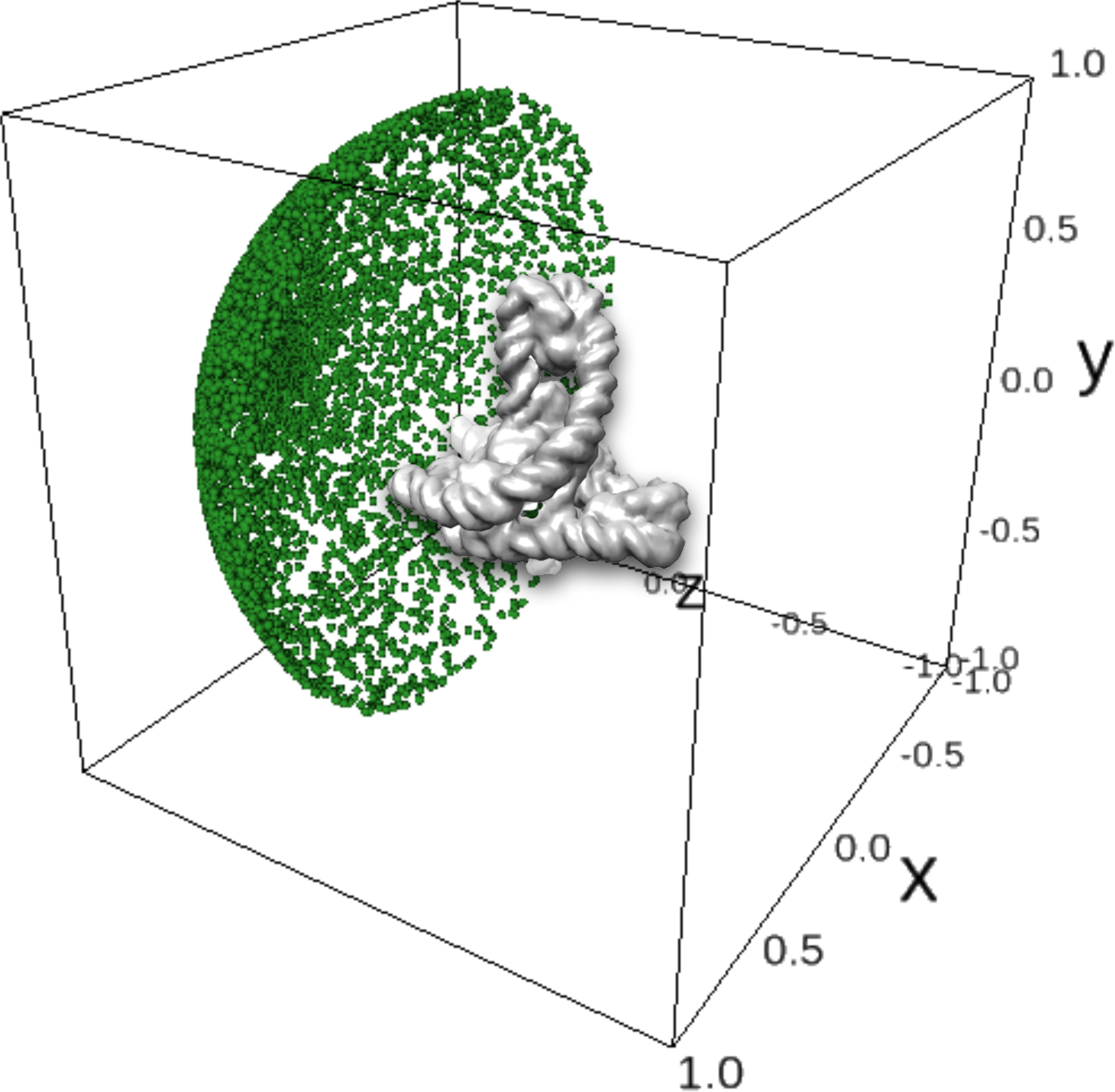}
            \hfill
            \includegraphics[height=2.5cm]{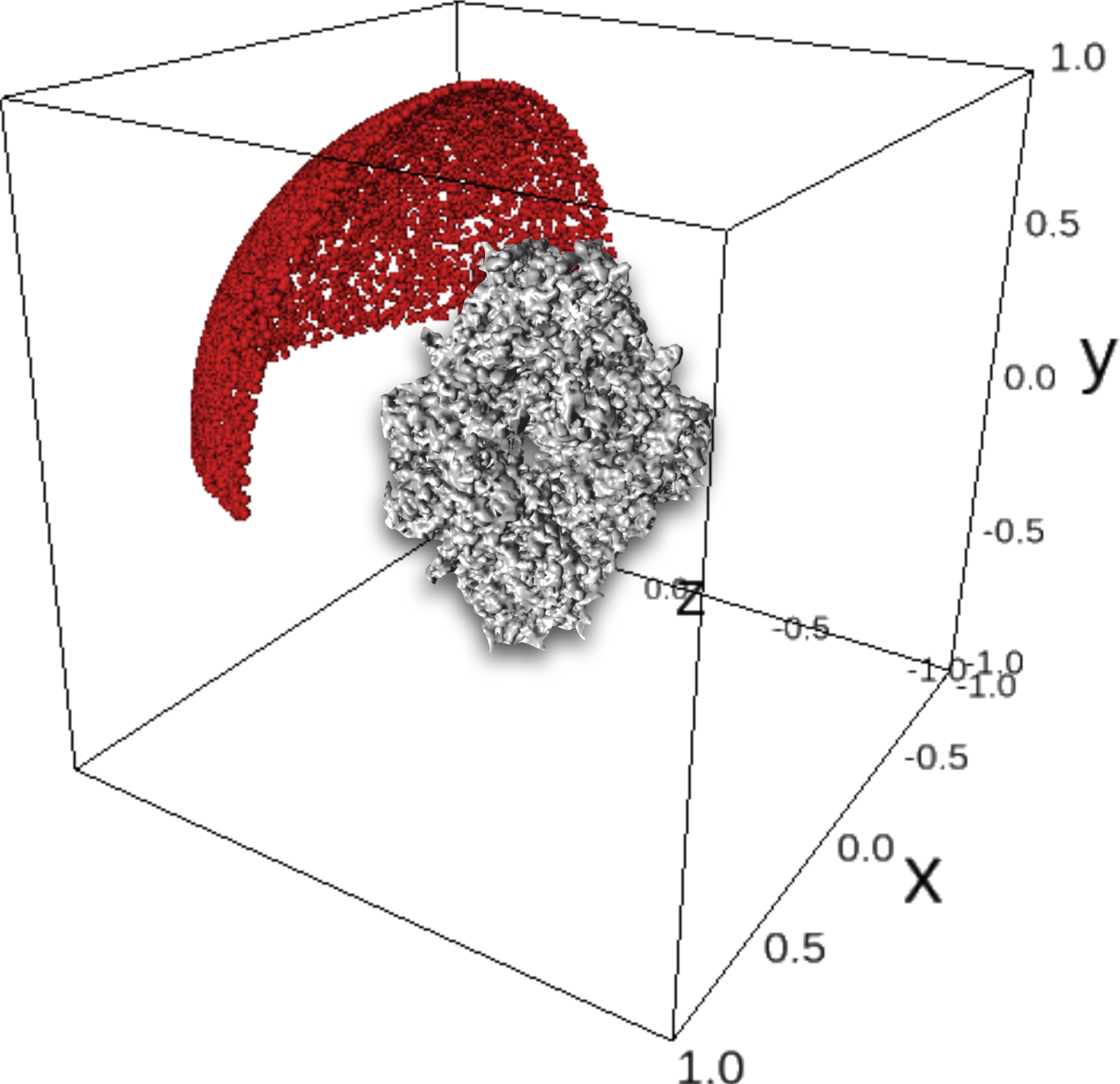}
            \caption{Sampled directions $(\theta_2, \theta_1)$.}%
            \label{fig:orientation-sampling:directions}
        \end{subfigure}
    \end{minipage}
    \hfill
    \begin{minipage}{.65\linewidth}
        \begin{subfigure}[b]{0.37\linewidth}
            \centering
            \includegraphics[height=2.4cm]{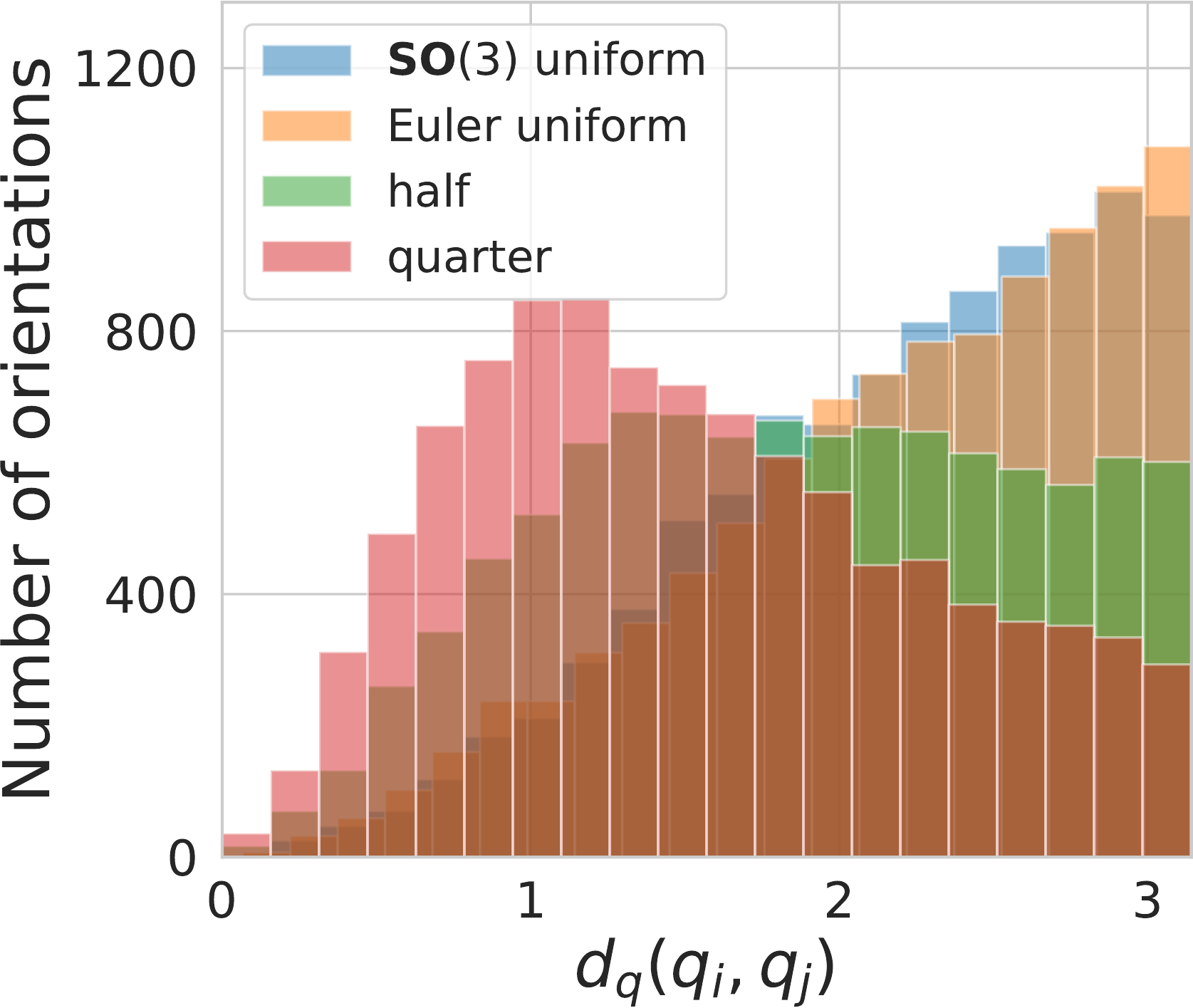}
            \caption{Distribution of distances $d_q$.}%
            \label{fig:orientation-sampling:distances}
            \vspace{0.2cm}
        \end{subfigure}
        \hfill
        \begin{subfigure}[b]{0.6\linewidth}
            \centering
            \includegraphics[height=2.4cm]{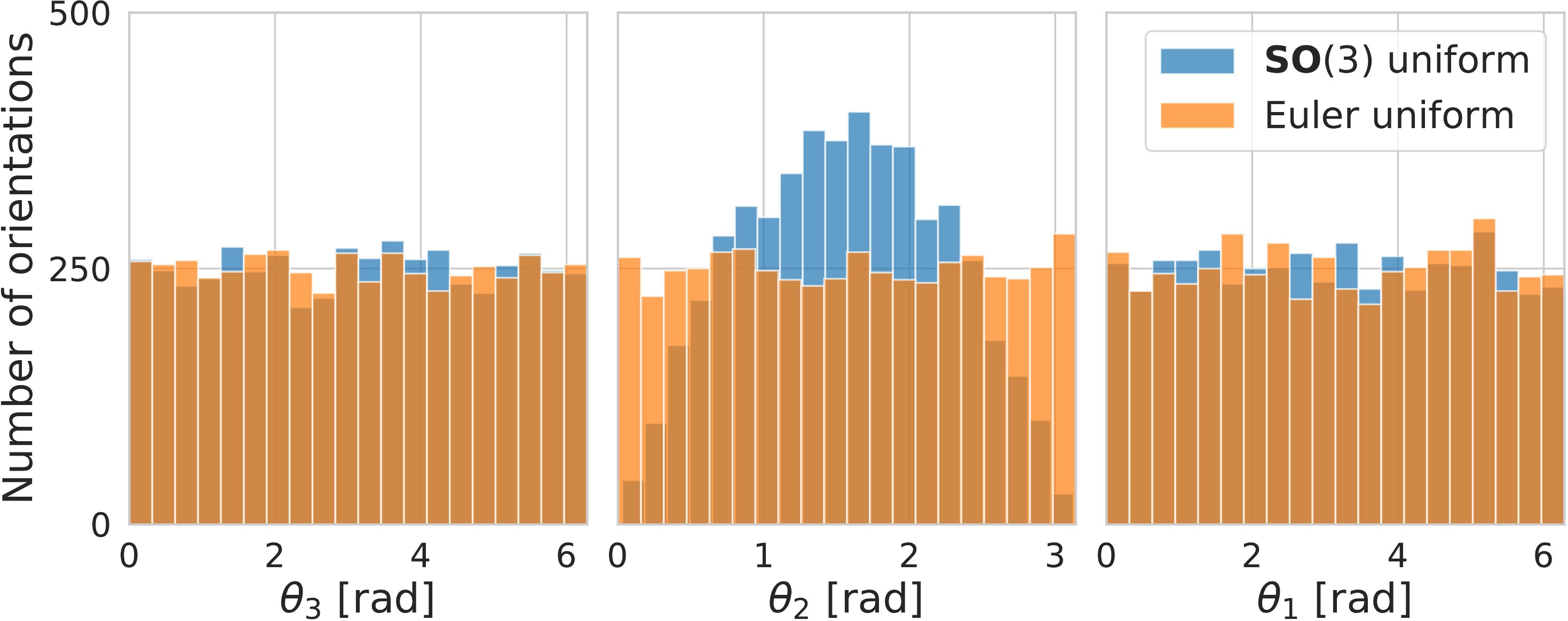}
            \caption{Distribution of Euler angles $\bth = (\theta_3,\theta_2,\theta_1)$.}%
            \label{fig:orientation-sampling:angles}
            \vspace{0.2cm}
        \end{subfigure}
        \begin{subfigure}[b]{0.97\linewidth}
            \centering
            \includegraphics[height=2.4cm]{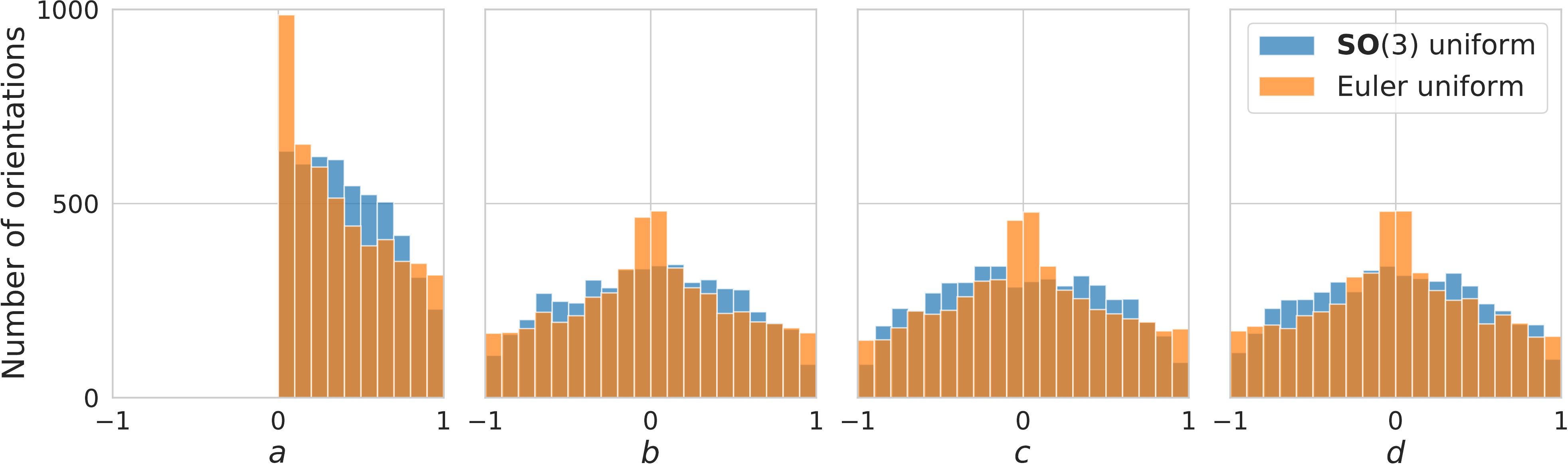}
            \caption{Distribution of quaternions $q = a + b\boldsymbol{i} + c\boldsymbol{j} + d\boldsymbol{k}$.}%
            \label{fig:orientation-sampling:quaternions}
        \end{subfigure}
    \end{minipage}
    \caption{%
        Sampling of orientations from four distributions:
        (blue) uniform on $\SO(3)$, (orange) uniform on Euler angles, (green) Euler uniform restricted to half the directions $(\theta_2, \theta_1) \in [0,\frac{\pi}{2}[ \, \times \, [0,2\pi[$, and (red) $\SO(3)$ uniform restricted to a quarter of the directions $(\theta_2, \theta_1) \in [0,\frac{\pi}{2}[ \, \times \, [0,\pi[$.
    }\label{fig:orientation-sampling}
\end{figure}

\figref{orientation-sampling} shows four distributions of orientations and the distributions of distances they induce.
As shorter distances are under-sampled, we uniformly resampled the distances to avoid biasing the training of our distance estimator.

\begin{figure}[ht!]
    \centering
    \begin{subfigure}[b]{0.48\linewidth}
        \centering
        \includegraphics[height=3cm]{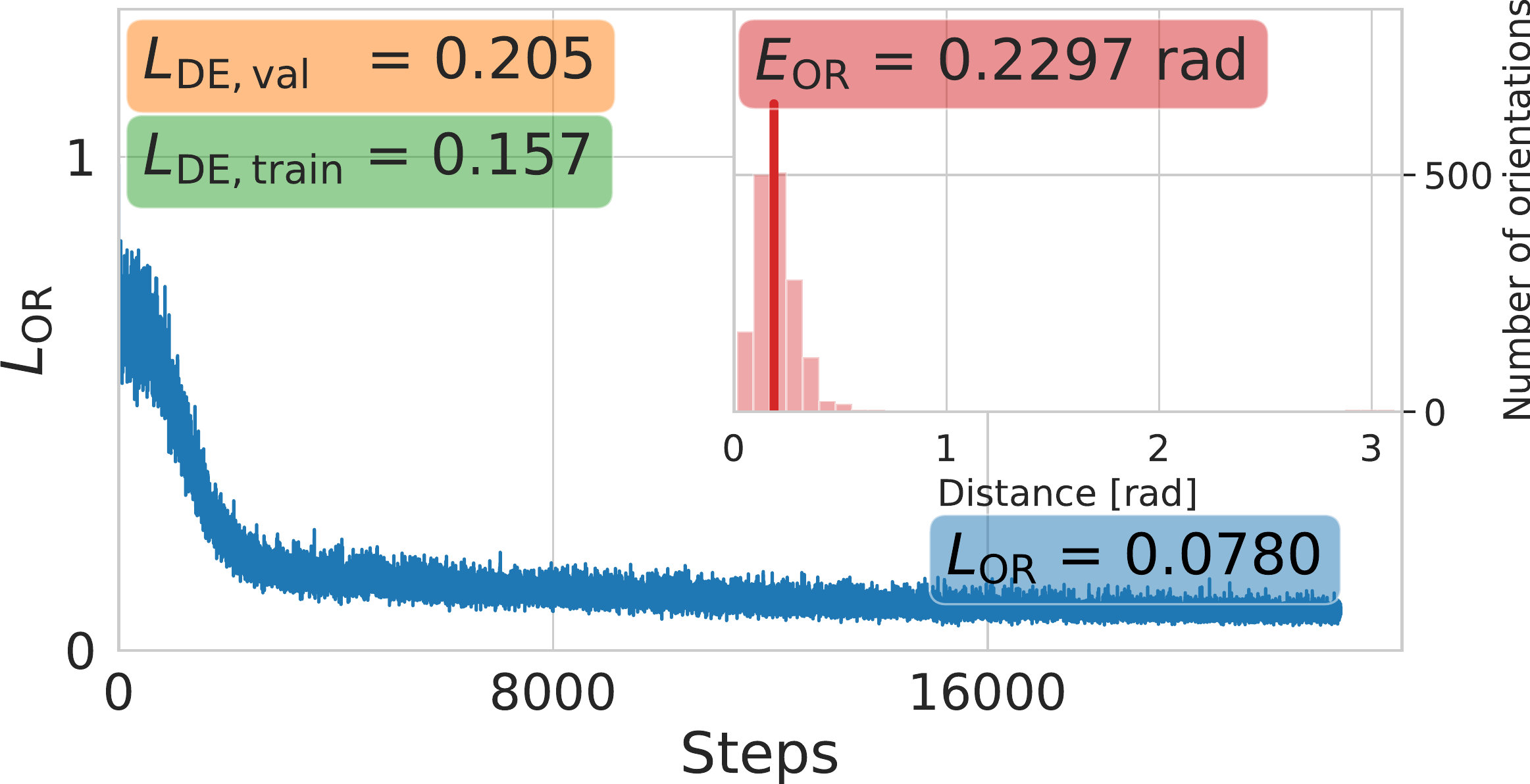}
        \caption{Distance learning and orientation recovery.}%
        \label{fig:nonuniform:recovery}
    \end{subfigure}
    \hfill
    \begin{subfigure}[b]{0.48\linewidth}
        \centering
        \includegraphics[height=3cm]{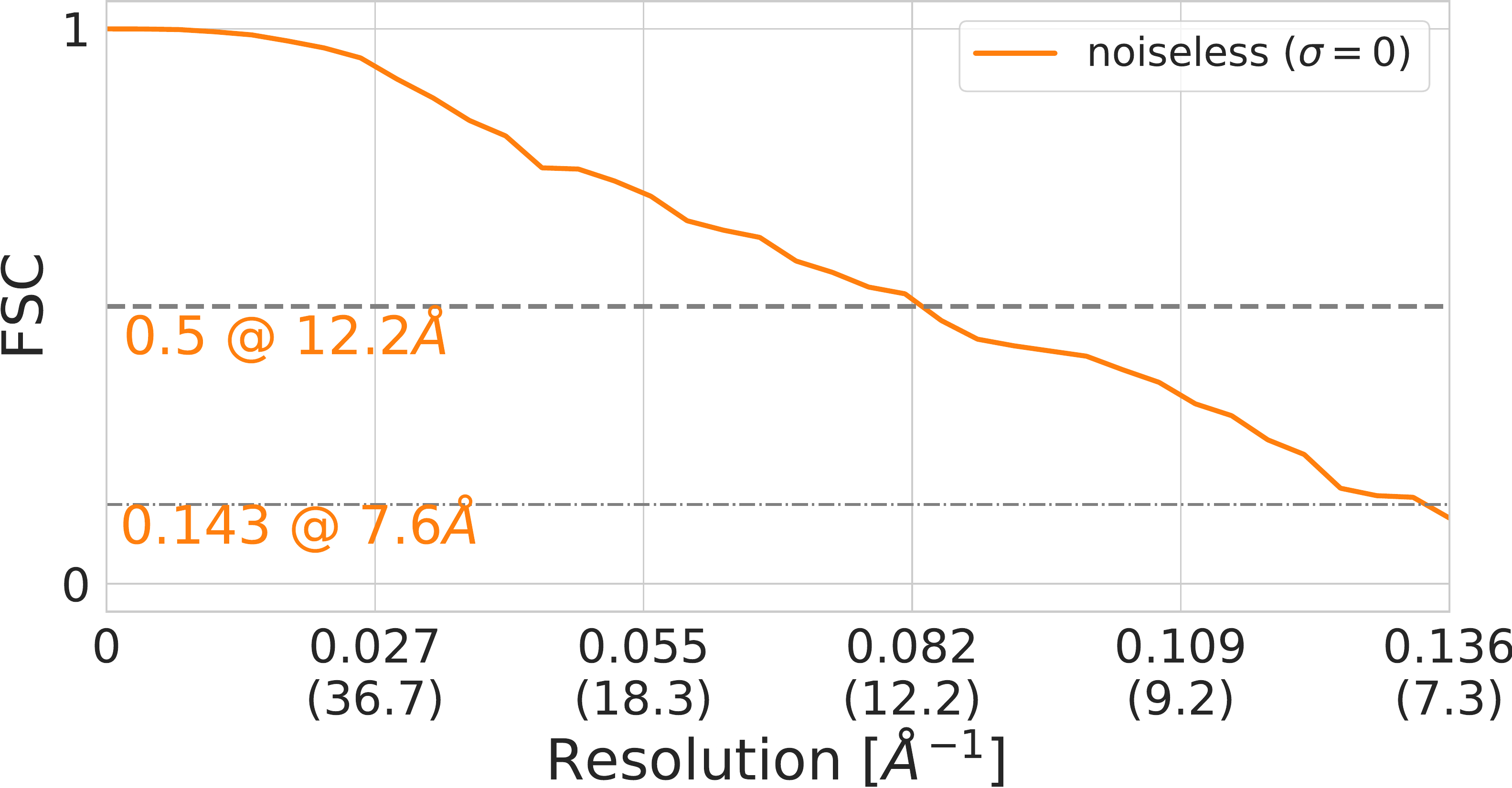}
        \caption{Fourier shell correlation (FSC) of the reconstructed density.}%
        \label{fig:nonuniform:reconstruction}
    \end{subfigure}
    \caption{%
        Orientation recovery and density reconstruction from noiseless projections of \texttt{5j0n} acquired from non-uniformly sampled orientations (uniformly sampled Euler angles, \figref{orientation-sampling} (orange)).
    }
\end{figure}

While we control the distributions of orientations and distances to facilitate distance learning, we cannot control them when recovering orientations of a given set of projections.
Comparing Figure~\ref{fig:nonuniform:recovery} with~\ref{fig:5j0n-noise0-orientation-recovery}, and~\ref{fig:nonuniform:reconstruction} with~\ref{fig:5j0n-reconstruction-fsc}, shows that the recovered orientations and the reconstructed density are barely affected by a non-uniform sampling of orientations---a condition that might happen in real cryo-EM acquisitions.

\section{Orientation recovery from exact distances}\label{apx:results:orientation-recovery:exact}

\begin{figure}[ht!]
    \begin{minipage}[t]{0.30\linewidth}
        \centering
        \includegraphics[height=3cm]{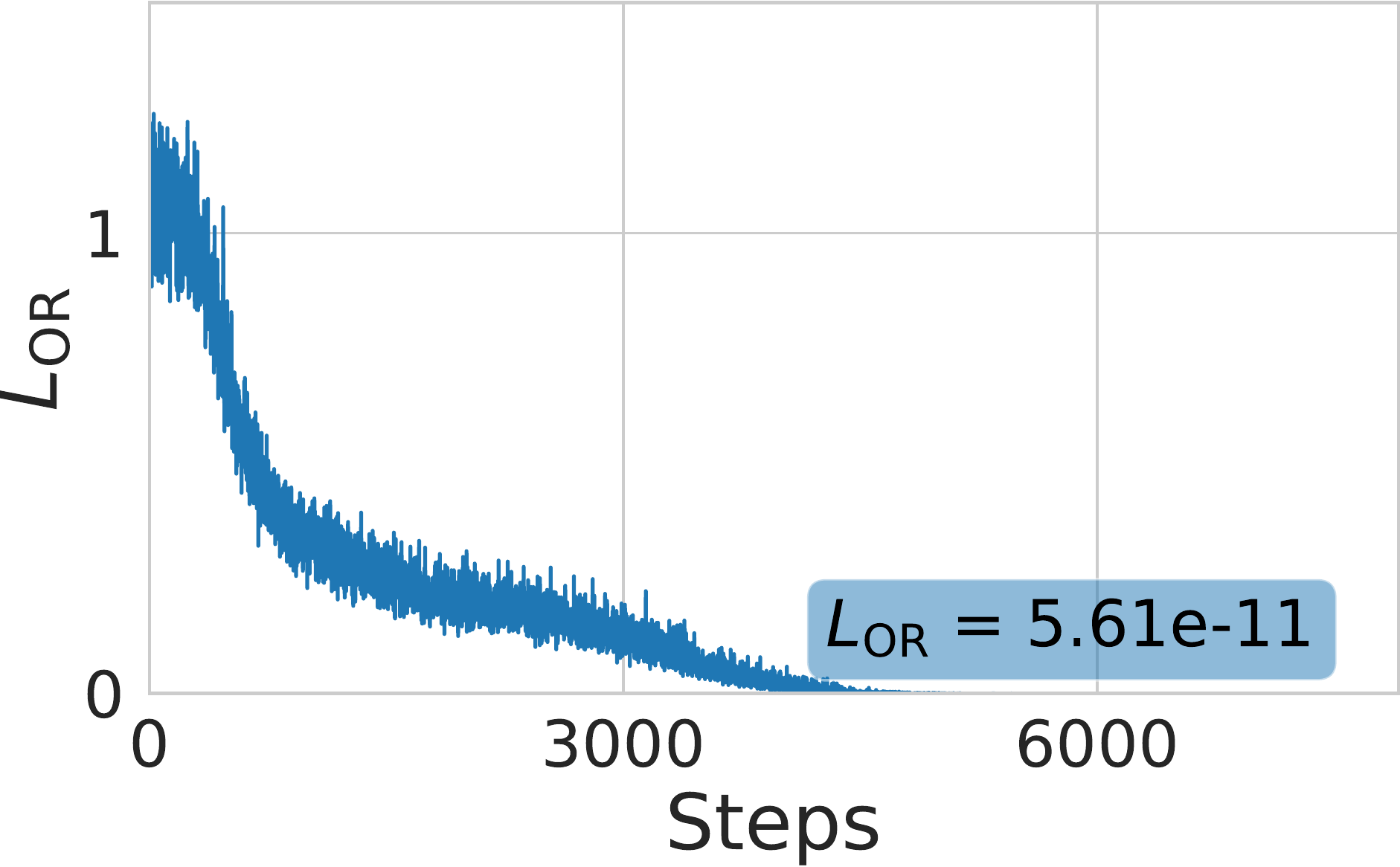}
        \caption{%
            Example of perfect orientation recovery (for \texttt{5a1a}).
            The loss $L_\text{OR}$ \eqnref{orientation-recovery} converges to zero when the distance estimation is perfect, \ie, $\widehat{d_p}(\p_i, \p_j) = d_q(q_i, q_j)$.
        }\label{fig:5j0n-orientation-recovery-loss}
    \end{minipage}
    \hfill
    \begin{minipage}[t]{0.66\linewidth}
        \begin{subfigure}[t]{0.49\linewidth}
            \centering
            \includegraphics[height=3cm]{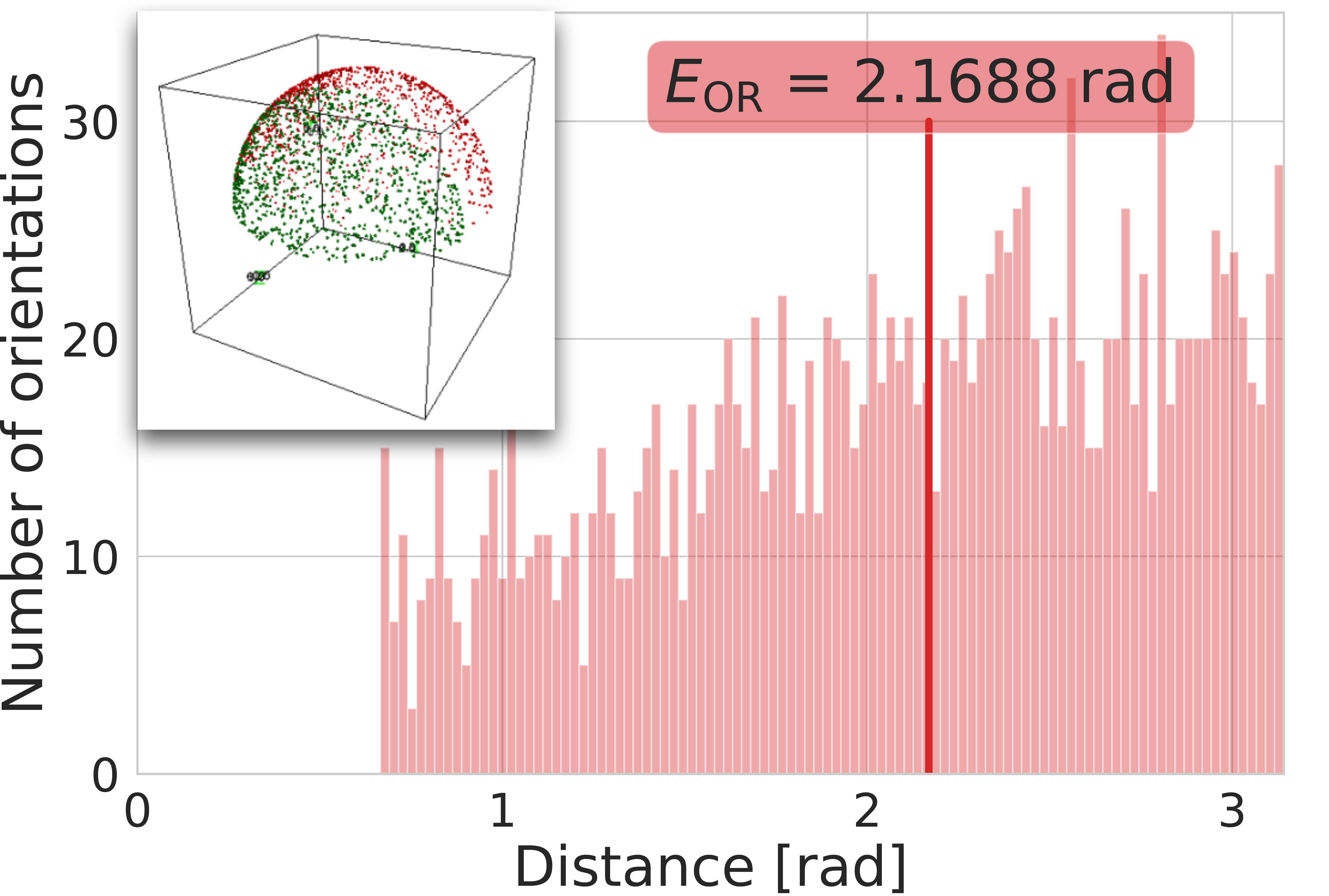}
            \caption{Orientations before alignment.}
        \end{subfigure}
        \hfill
        \begin{subfigure}[t]{0.49\linewidth}
            \centering
            \includegraphics[height=3cm]{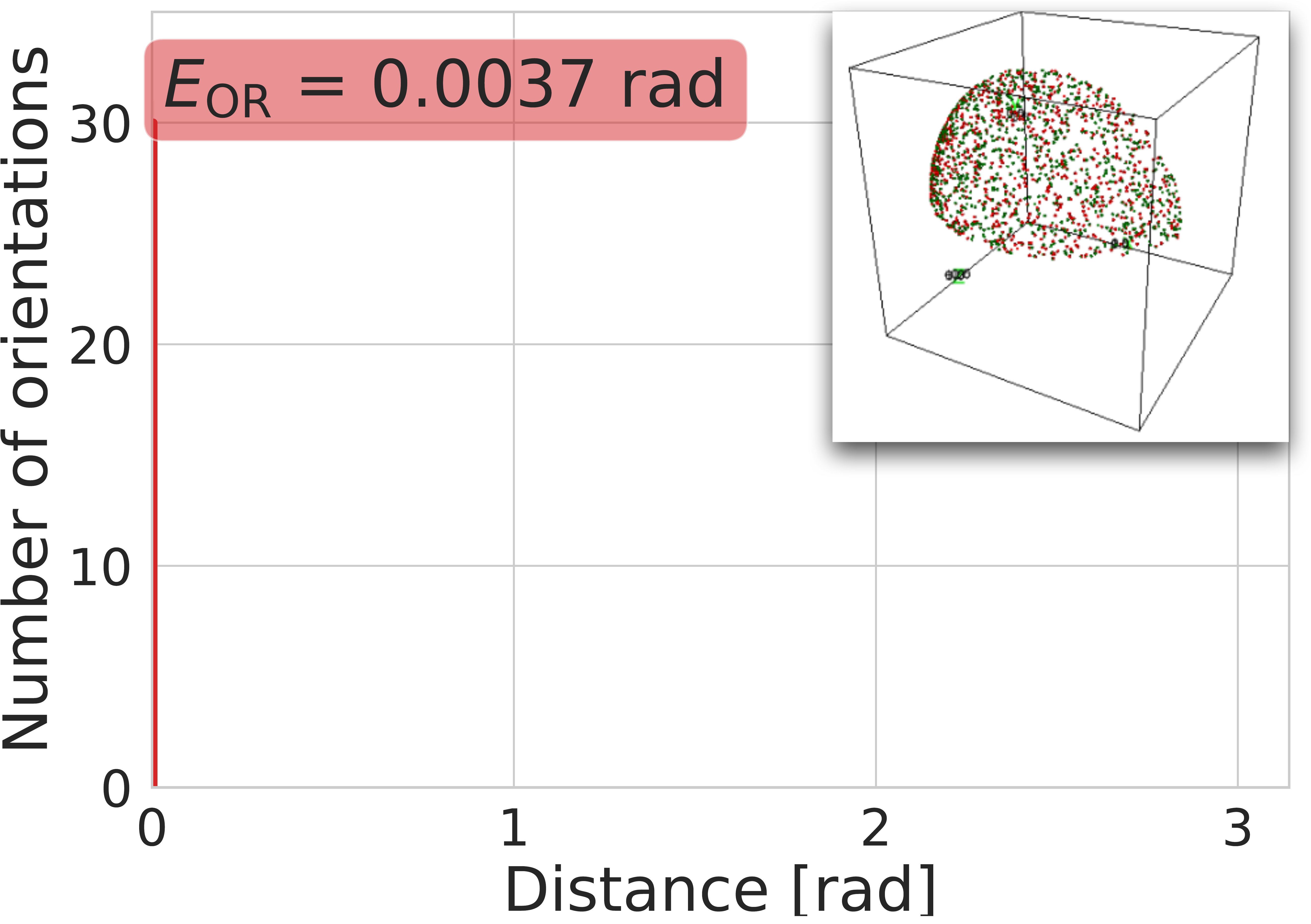}
            \caption{Orientations after alignment.}
        \end{subfigure}
        \caption{%
            Example of perfect alignment~\eqnref{orientation-recovery-error} after a perfect recovery~\eqnref{orientation-recovery}.
            The red histogram shows the errors (a) $\{d_q(q_i, \widehat{q_i})\}$ and (b) $\{d_q(q_i, \T \widehat{q_i})\}$, with the mean $E_\text{OR}$ highlighted.
            True (green) and recovered (green) directions are shown in the insert.
            While both colors are seen in (b), they are superimposed.
        }\label{fig:5j0n-aa-loss-perfect-distances}
    \end{minipage}
\end{figure}

To verify that the lack of a convexity guarantee for \eqnref{orientation-recovery} and the sampling of the sum are non-issues in practice, we attempted orientation recovery under exact distance estimation $\widehat{d_p}(\p_i, \p_j) = d_q(q_i, q_j)$.
Orientations were perfectly recovered;
\figref{5j0n-orientation-recovery-loss} shows the convergence of $L_\text{OR}$ to zero.
\figref{5j0n-aa-loss-perfect-distances} shows how~\eqnref{orientation-recovery-error} could then perfectly align the recovered and true orientations---leading to $E_\text{OR} = 0$.
It illustrates how alignment is necessary to evaluate the performance of orientation recovery.

\section{Euclidean distance between projections}\label{apx:results:distance-estimation:euclidean}

\begin{figure}
    \begin{minipage}[t]{0.99\linewidth}
        \begin{subfigure}[t]{0.33\textwidth}
            \centering
            \includegraphics[height=3cm]{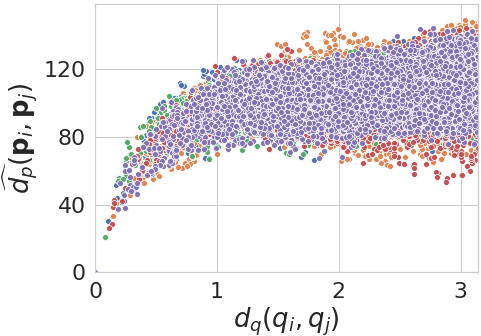}
            \caption{Full direction coverage on \texttt{5j0n}.}%
            \label{fig:euclidean-not-robust:5j0n-full}
        \end{subfigure}
        \hfill
        \begin{subfigure}[t]{0.33\textwidth}
            \centering
            \includegraphics[height=3cm]{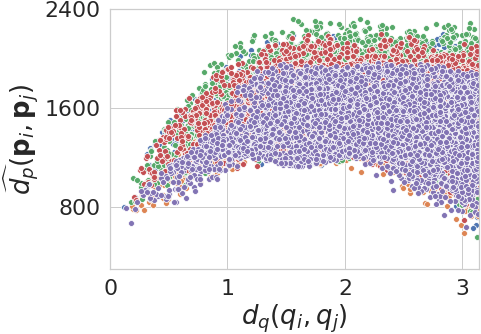}
            \caption{Full direction coverage on \texttt{5a1a}.}%
            \label{fig:euclidean-not-robust:5a1a-full}
        \end{subfigure}
        \hfill
        \begin{subfigure}[t]{0.33\textwidth}
            \centering
            \includegraphics[height=3cm]{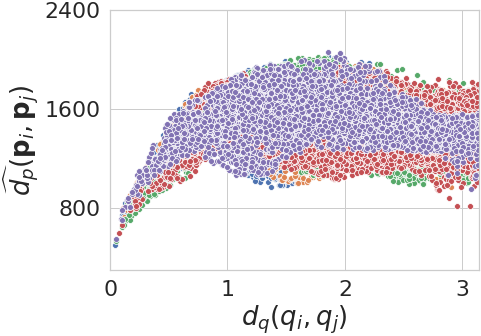}
            \caption{Quarter direction coverage on \texttt{5a1a}.}%
            \label{fig:euclidean-not-robust:5a1a-quarter}
        \end{subfigure}
        \caption{%
            Euclidean distance between projections $\widehat{d_p}(\p_i, \p_j) = \| \p_i - \p_j \|_2$ versus their actual relative orientation $d_q(q_i, q_j)$.
            We randomly selected $5$ projections from $P = 5,000$: each color represents the distances between one of those and all the others.
        }\label{fig:euclidean-not-robust}
    \end{minipage}
\end{figure}

We evaluate $\widehat{d_p}(\p_i, \p_j) = \| \p_i - \p_j \|_2$ (\ie, the Euclidean distance) as a baseline distance estimator.
\figref{euclidean-not-robust} shows the relationship between $\widehat{d_p}$ and $d_q$.
Two main observations can be made from this experiment.
First, as suspected, $\widehat{d_p}$ fails to be a consistent predictor of $d_q$, even in the simple imaging conditions considered here (no noise, no shift, no PSF).
In particular, the larger the orientation distance $d_q$, the poorer the predictive ability of $\widehat{d_p}$ (the plot plateaus).
Second, because \texttt{5a1a} has D2 symmetries, two projections might be identical while not having been acquired from the same orientation.
Restricting directions to a quarter captures only one of four identical projections, solving the issue.

\section{SiameseNN: feature distance and embedding dimension}\label{apx:siamese:feature-distance-and-embedding-dimension}

\begin{figure}[ht]
    \centering
    \begin{subfigure}[t]{0.44\linewidth}
        \centering
        \includegraphics[height=3.5cm]{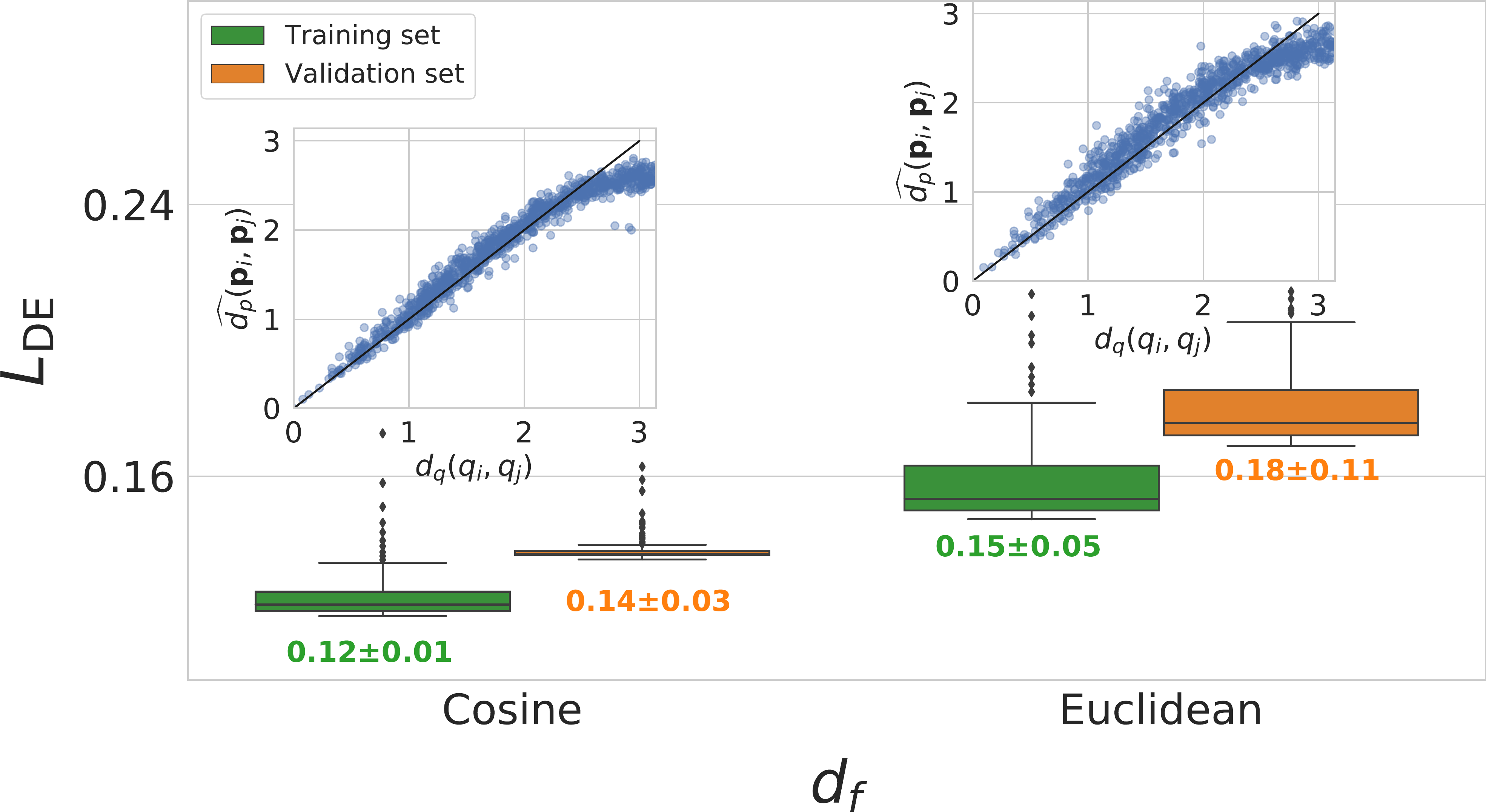}
        \caption{%
            Performance w.r.t.\ the feature distance $d_f$.
        }\label{fig:geo-eucl-mlp}
    \end{subfigure}
    \hfill
    \begin{subfigure}[t]{0.51\linewidth}
        \centering
        \includegraphics[height=3.5cm]{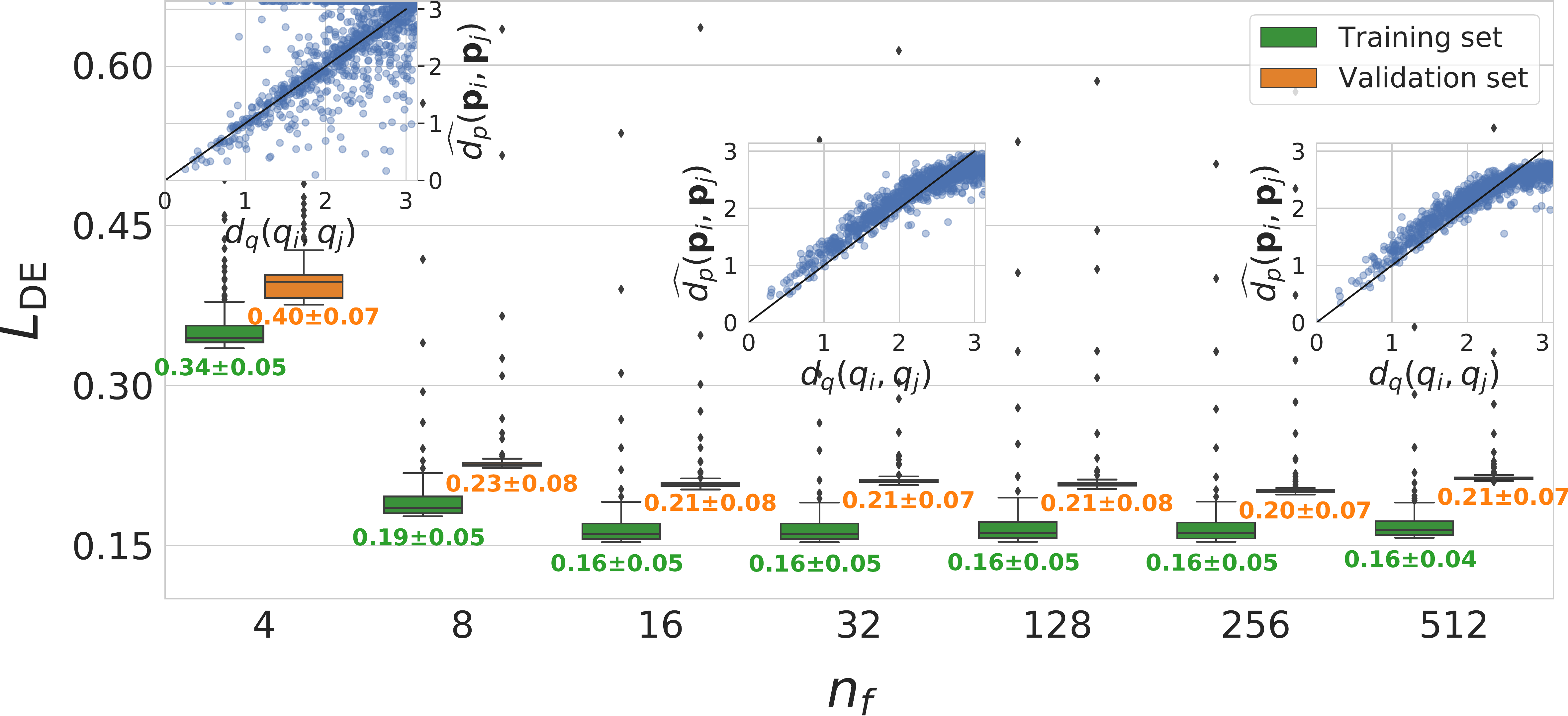}
        \caption{%
            Performance w.r.t.\ the embedding dimensionality $n_f$.
        }\label{fig:4d-vs-256d-de}
    \end{subfigure}
        \caption{%
            Performance of our distance estimator $\widehat{d_p}$ w.r.t.\ two design choices.
            The box plots show the distance learning loss $L_\text{DE}$ \eqnref{distance-learning}.
            The inserted plots show the relationship between $d_p(\p_i, \p_j) = d_f(\mathcal{G}_w(\p_i), \mathcal{G}_w(\p_j))$ and $d_q(q_i, q_j)$ on $1,000$ pairs sampled from \texttt{5j0n}.
        }
\end{figure}

There are multiple options for a distance function $d_f$ between two features $\mathbf{f}_i = \mathcal{G}_w(\p_i) \in \R^{n_f}$.
\figref{geo-eucl-mlp} compares the use of the Euclidean distance $d_f(\f_i, \f_j) = \Vert \f_i - \f_j \Vert_2$ and the cosine distance $d_f(\f_i, \f_j) = 2 \arccos \left( \frac{\langle \f_i, \f_j \rangle}{\lVert \f_i \rVert \lVert \f_j \rVert} \right)$. The cosine distance results in a lower $L_\text{DE}$, which makes $\widehat{d_p}$ a better estimator of $d_q$.
This superiority of the cosine distance is likely due to its capacity to model the elliptic geometry of $\SO(3)$, a feat the Euclidean distance does not achieve, the Euclidean space being neither periodic nor curved.

\figref{4d-vs-256d-de} shows the performance of our distance estimator $\widehat{d_p}$ depending on the size $n_f$ of the feature space.
It clearly indicates that a space of $n_f=4$ dimensions is insufficient to represent the variability of projections.
That is a motivation to embed the projections in a space of higher dimensions that can represent more variations than the orientation, and can abstract that variation by solely considering the distances between the embedded projections $\f_i = \G_w(\p_i)$.
While our choice of $n_f=512$ might be overkill ($n_f=16$ seems sufficient), it is not penalizing.

\newpage
\section{Convolutional neural network architecture}\label{apx:siamese-architecture}

\begin{figure}[ht!]
    \centering
    \begin{subfigure}[t]{1.0\linewidth}
        \includegraphics[width=\linewidth]{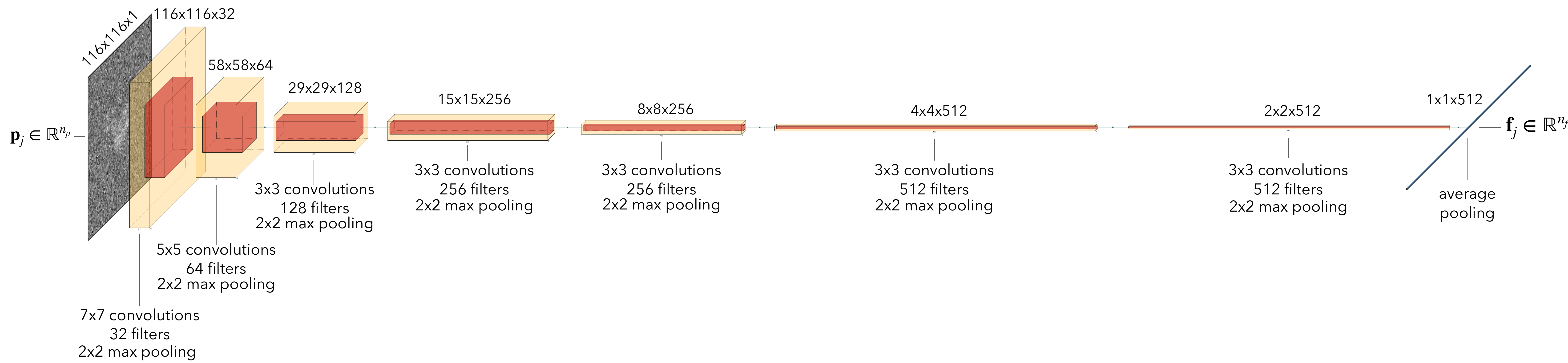}
    \end{subfigure}
    \caption{%
        Architecture of $\mathcal{G}_w$, the convolutional neural network that extracts feature vectors $\f_j = \mathcal{G}_w(\p_j) \in \R^{n_f}$ from projections $\p_j \in \R^{n_p}$.
        While $n_f=512$ and $n_p=116 \times 116$ in our experiments, $\mathcal{G}_w$ can accommodate any image size thanks to the global average pooling layer.
    }\label{fig:de-architecture}
\end{figure}

\end{document}